\pdfoutput=1
\documentclass[review]{elsarticle}
\usepackage{subfigure}
\usepackage{longtable}
\usepackage{algorithm, algorithmic}
\usepackage{bm}
\usepackage{booktabs}
\usepackage{multirow}
\usepackage{geometry}

\usepackage{lineno,hyperref}
\modulolinenumbers[5]

\journal{arXiv}

\begin{document}

\begin{frontmatter}

\title{Survival Analysis of the Compressor Station Based on Hawkes Process with Weibull Base Intensity}


\author[mymainaddress]{Lu-ning Zhang}
\ead{luning_zhang@foxmail.com}

\author[mymainaddress]{Jian-wei Liu\corref{mycorrespondingauthor}}
\cortext[mycorrespondingauthor]{Corresponding author}
\ead{liujw@cup.edu.cn}

\author[mymainaddress]{Xin Zuo}

\address[mymainaddress]{Department of Automation, College of Information Science and Engineering,
China University of Petroleum , Beijing, Beijing, China}

\begin{abstract}
In this paper, we use the Hawkes process to model the sequence of
failure, i.e., events of compressor station and conduct survival
analysis on various failure events of the compressor station.
However, until now, nearly all relevant literatures of the Hawkes
point processes assume that the base intensity of the conditional
intensity function is time-invariant. This assumption is apparently
too harsh to be verified. For example, in the practical application,
including financial analysis, reliability analysis, survival
analysis and social network analysis, the base intensity of the
truth conditional intensity function is very likely to be
time-varying. The constant base intensity will not reflect the base
probability of the failure occurring over time. Thus, in order to
solve this problem, in this paper, we propose a new time-varying
base intensity, for example, which is from Weibull distribution. First, we
introduce the base intensity from the Weibull distribution, and then
we propose an effective learning algorithm by maximum likelihood
estimator. Experiments on the constant base intensity synthetic
data, time-varying base intensity synthetic data, and real-world
data show that our method can learn the triggering patterns of the
Hawkes processes and the time-varying base intensity simultaneously
and robustly. Experiments on the real-world data reveal the Granger
causality of different kinds of failures and the base probability of
failure varying over time.
\end{abstract}


\end{frontmatter}

\section{Introduction}

Learning point processes, especially the Hawkes process from
irregular and asynchronous sequential data observed in continuous
time is a challenging task. Point processes can be applied to many
fields, such as Ad serving, disease prediction, and TV shows
recommendation. All of these sequence data can be modeled by a point
process, i.e., we can use the Hawkes processes to model the
triggering patterns between the different types of event, i.e., the
TV shows user was watching, the ad users clicked and patient's
disease which the subject is getting sick. Similar to these ideas,
we want to understand the trigger mode between various faults in the
compressor station, so we applied the point process model in the
survival analysis of the compressor station.

Due to the research of the causality relationship between different
types of event of point process \cite{didelez2008graphical}, there
are a lot of research focusing on Hawkes processes, which is
proposed by Hawkes \cite{hawkes1971spectra}, and widely applied in
many fields. Recently, an effective method is proposed to learn
multi-dimensional Hawkes processes \cite{hawkes1971spectra} and
Granger causality of different kinds of event by learning the impact
function of Hawkes processes and the causality relationship with
different event types from the sequential data
\cite{xu2016learning}, from the perspective of the graphical model,
learning causality relationship with different event types is
equivalent to learning Granger causality graph. In Granger causality
graph, the line with arrow connecting two types of event nodes
indicates that the event of the dimension corresponding to the
destination node is dependent on the historical events of the
dimension corresponding to the source node.

\cite{xu2016learning} proposed an effective method to learn Hawkes
processes and then use them to deduce the Granger causality hiding
in the impact function, and applied it into IPTV data, to reflect
the trigger mode of users watching preference. However, before this,
learning Granger causality for general multi-dimensional point
processes with irregular and asynchronous event sequence is hard to
accomplish. Existing works mainly focus learning Granger causality
from time series \cite{arnold2007temporal, eichler2012graphical,
basu2015network}. \cite{han2013transition} proposed a vector
auto-regressive (VAR) model to learn Granger causality from discrete
time-lagged variables. However, learn Granger causality from point
processes is more difficult, because the event sequence of point
processes is continuous evolving over time and doesn't have fixed
time-lag. So, it was hard to come up with an effective method to
capture the Granger causality between different types of event.
\cite{lian2015multitask} proposed a method which learns Granger
causality by constructing features from history and selecting them.
However, this method relies heavily on feature constructing
processes, may resulting in poor robustness.

In this paper, we apply the Hawkes process to the failure/survival
analysis of the compressor station. First, we utilize the Hawkes
process to analyze the causal relationship between different types
of failures. We also devise the approach to get the impact of the
possibility of other failures after a certain failure occurs. While
analyzing the causal relationship between failures of the compressor
station, we also want to identify the base probability of each kind
of failures occurred over time. However, this idea is difficult to
implement based on the existing Hawkes process model. The reason is
that nearly all of the Hawkes processes models researcher proposed,
assume the base intensity is constant. This assumption has a lot of
restriction, for instance, in disease prediction, the base intensity
of patient get disease shouldn't be constant, it must be
time-varying, as the patient gets older, the risk of he getting sick
should be bigger, in the field of TV shows recommendation, the
attractive of TV shows is also varying over time, should not be a
constant.

On the other hand, failure event sequences have some characteristics that other event sequences do not have, figure 1 depicts these characteristics of failure event sequences.
During the normal operation of the gas station, the probability of failure is relatively low, however, once a failure occurs, the likelihood of derivative failures will increase relatively.
Thus, on the time axis, the distribution of failure events presents a sparse-clustering feature. We call a sequence of events with such characteristics a sparse-clustering sequence, the first failure in a clustering, we called as source failure. When the device is working normally, the failure occurs sparsely, and after one failure occurs, due to the trigger mode between the source failure and derivative failures, there is a phenomenon of clustering between them, and we assume that the source failure did not receive the impact of the previous failure history.

Thus, in order to better guide production, we not only learning the trigger patterns between different kinds of failures, but also have to figures out the probability of occurrence of source failures changes with time, in this condition ,we also have to learn the time base intensity of failures.

\begin{figure}[h]
\centering
\includegraphics[width=1\textwidth]{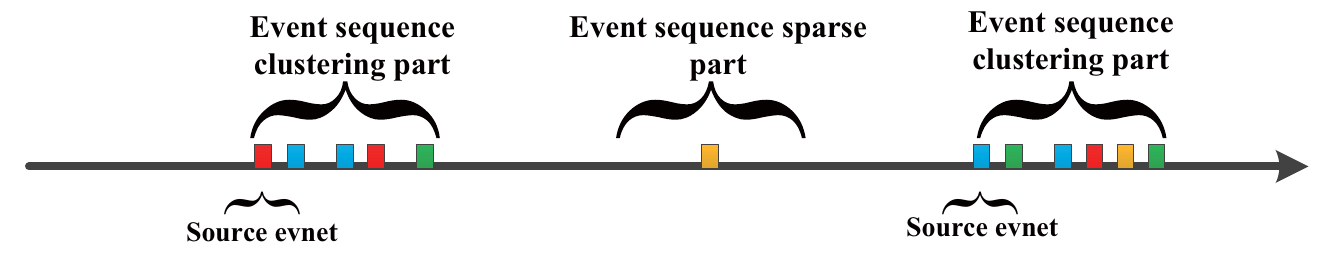}
\caption{The schematic diagram of sparse-clustering sequence}
\label{111}
\end{figure}

In order to solve these problems, we introduce a kind of time-varying
base intensity, i.e., the base intensity from Weibull distribution,
which is widely used in many fields, such as survival analysis,
reliability engineering and weather forecasting, can express the
trend of the base intensity over time. The introduction of Weibull
base intensity brings a new parameter   into the model. We propose a
new effective algorithm to learn it. After estimating the parameter
, we can obtain the trend of the instantaneous occurrence rate of
all kinds of failure, which can help the compressors to develop a
plan of equipment overhaul and maintenance to reduce the occurrence
of compressor failures and improve production efficiency, which can
bring greater economic value and safety value.

In Section 2, we discuss related basic concepts, about point
process, Weibull distribution and exponential distribution. In
section 3, we theoretically deduced the proposed model and proposed
a learning algorithm for model parameters. In section 4, we
validated our model on two different synthetic datasets,
demonstrating that our model is valid for both time-varying base
intensity and constant base intensity, then, we validated our model
on real-world data, it proves that our model is more effective on
the real-world data, and obtains the base trend with time and the
causal relationship of all kinds of failures.

\section{Related work}

\paragraph{Survival analysis} Survival analysis is a branch of statistics for
analyzing when one or more events happen, such as death of an
infectious disease patients and failure in mechanical systems. In
engineering, it is called as reliability theory or reliability
analysis, in economics, it is called duration analysis or duration
modeling, and in sociology it is called event history
analysis.\cite{duchateau2007frailty} systematically introduce the
survival analysis and propose the new frailty model. In the field of
reliability analysis, there have been many research results, such as
\cite{pate1984fault}, \cite{pate2004limitations} and
\cite{aslett2015bayesian}.We will adopt the relevant knowledge of
survival analysis in engineering field to analyze the reliability of
the compressor station subsystems.

\paragraph{Weibull distribution} Weibull distribution is first identified by
\cite{frechet1927loi} and first apply to describe a particle size
distribution in \cite{rosin1933laws}. However, Swedish mathematician
Waloddi Weibull, describe it in detail in
\cite{weibull1951statistical}, so this distribution is named as
Weibull distribution. Weibull distribution is widely applied in so
many areas, such as life time prediction \cite{ali2015accurate},
reliability analysis \cite{bain2017statistical}, survival analysis
\cite{cox2018analysis} , weather forecasting and the wind power
industry to describe wind speed distributions \cite{mohammadi2016assessing}.

\paragraph{Point processes} The temporal point process is an event sequence randomly located on
time space where the knowledge related to survival analysis can be
applied. For instance, hazard function in survival analysis is
equivalent to the conditional intensity function in temporal point
process. Thus, we will model the failure sequences of compressor
station by Hawkes processes for survival analysis. First, we need to
introduce the most basic point process model, i.e., Poisson process.
Poisson processes \cite{vere2003introduction} are the simplest point
process model, the difference between the other point processes and
Poisson processes is that the current event of Poisson process is
independent of the historical events. Poisson process has following
property:

\[{\lambda _c}(t) = {h_c}(t)\]

Generally speaking, Poisson process can be divided into two
categories, Poisson process with time-varying ${h_c}(t)$, commonly
called the nonhomogeneous or inhomogeneous Poisson process; Poisson
process with constant ${h_c}(t)$ commonly called homogeneous Poisson
process.

However, pure Weibull distribution, survival analysis or Poisson
process do not consider the trigger mode between various faults, so
we need to combine these theories with the Hawkes process and
Granger causality, then analyze the sequences of faults more
effectively and comprehensively. Hawkes processes
\cite{hawkes1971spectra} are proposed to model the complicated event
sequences which events occurred in the past will affect the
occurrence of future events. Hawkes processes is an important kind
of mutually exciting point processes which is applied to many
practical fields, e.g., financial analysis
\cite{embrechts2011multivariate, bacry2015hawkes}, social network
modeling \cite{zhou2013learninga}, and seismic
analysis\cite{daley2007introduction}.

The multidimensional Hawkes process is described by its conditional
intensity function (also called as hazard function in survival
analysis) which has following form:

\[\begin{array}{l}
{{\rm{\lambda }}_c}(t) = {h_c}(t) + \sum\nolimits_{c' = 1}^C {\int_0^t {{\phi _{cc'}}(s)d{N_{c'}}(t - s)} } \\
{\rm{       }} = {h_c}(t) + \sum\nolimits_{c' = 1}^C {\int_0^t
{{\phi _{cc'}}(t - s)d{N_{c'}}(s)} }
\end{array}\]

${h_c}(t)$ is the base intensity, which is independent of the
history information. Generally speaking, the Hawkes processes
research introduced in this section usually assumed that ${h_c}(t) =
\mu$ the base intensity is constant and
time-invariant.\cite{laub2015hawkes} gives a detailed overview of
the Hawkes process, introduce the mathematical model of the Hawkes
process, parameter estimation methods, simulation methods and so on.
in the last few years, the most of research works which is related
to Hawkes processes use predefined impact function with fixed
parameters, for example , the power-law functions is used in
\cite{zhao2015seismic}, and the exponential function is used in
\cite{zhou2013learninga, hall2016tracking, farajtabar2014shaping,
yan2015machine, lewis2011nonparametric} however, these methods
oversimplify the impact function, thus may not learn the Hawkes
processes well. There are also some models that have learned the
impact function. The main idea of these approaches is to use a
non-parametric model for enhancing the flexibility. The 1-D Hawkes
processes with nonparametric model is first proposed in
\cite{lewis2011nonparametric} ,which is based on ordinary
differential equation, then this method is extended to
multi-dimensional case in \cite{zhou2013learningb, luo2015multi}.
\cite{bacry2012non} estimates the nonparametric Hawkes processes via
solving the Wiener-Hopf equation. \cite{reynaud2010adaptive,
hansen2015lasso, eichler2017graphical} use the contrast
function-based estimation method, minimizes the estimation error of
the conditional intensity function and transfer the problem to a
Least-Squares(LS) problem. \cite{du2012learning,
lemonnier2014nonparametric} decompose impact functions into basis
functions to avoid discretization.

The Gaussian process-based methods have been reported to
successfully estimate more general point processes
\cite{lian2015multitask,adams2009tractable,lloyd2015variational,
samo2015scalable}. However, these methods are too complicated to
learn so many parameters, resulting in loss of model
performance.\cite{eichler2017graphical} first demonstrate the
relationship between the impact function of the Hawkes process and
Granger causality.\cite{xu2016learning} propose a nonparametric
model to learn the Hawkes processes by an effective learning
algorithm combining a maximum likelihood estimator (MLE) with a
sparse-group-lasso (SGL) regularizer, this method can learn the
impact function and Granger causality robustly, then they propose an
adaptive procedure to select basis functions. These Hawkes-based
methods all set the base intensity to a constant, introducing
unnecessary restrictions for model assumption on the survival
analysis of faults. \cite{mei2017neural, xiao2017modeling} propose
to model streams of discrete events in continuous time, creatively
construct a neural self-modulating multivariate point process, which
is called the neural Hawkes process, however, these two methods
based on recurrent neural networks are black box models, which can
be inconvenient to reveal the trend of the possibility of failure
events themselves, and the trigger mode between faults.

And to our best knowledge, no one has ever introduced Weibull base
intensity into Hawkes process, we will utilize the Weibull
distribution for failure survival analysis to evaluate the system
reliability of the compressor station, and our research will inject
new impetus into the Hawkes process.

\paragraph{Granger causality} There are a lot of research efforts in Granger causality of point processes \cite{meek2014toward},
for more general random processes, \cite{chwialkowski2014kernel} come up with a kernel independence test method.
\cite{ahmed2009recovering} applies lasso and its variants to reveal the inner-structure of nodes which is a common way to learn Granger causality.
 \cite{gunawardana2011model} proposes a model can identify temporal dependencies between different types of event.
  \cite{basu2015network} takes inherent grouping structure into considered when leaning Granger causality from discrete transition process.
   Based on parametric cascade generative process, \cite{daneshmand2014estimating} puts forward inference model for continuous time
   diffusion network to learn Granger causality.\cite{didelez2008graphical} proposes a theorem that reflects the relationship between
   the impact function and Granger causality, the most effective method is learning Granger causality from the impact function in the
    conditional intensity of point processes. And \cite{eichler2017graphical} specializes the work for Hawkes processes and reveals the
    relationship between impact function and Granger causality. \cite{xu2016learning} comes up with a nonparametric model to
    learn the Hawkes processes, the Granger causality graph and the triggering patterns of the Hawkes processes simultaneously.
    Since these methods does not consider the effects of time-varying base intensity, the results of causal analysis has received a negative impact,
     resulting in the Granger relationship obtained by these methods is not the most sparse and most accurate.
     In this paper, we will utilize a similar algorithm to analyze the causal relationship between the failures of the various subsystems
     in the gas station so as to prevent the occurrence of secondary failures.

\section{Basic Concepts}

\subsection{Point process}

\paragraph{Definition 1 (Counting process)}A counting processes is a stochastic processes $(N(t):t \ge
{T_b})$, ${T_b}$ is the beginning of the observation window. A
counting process is almost surely finite, and is a right-continuous
step function which the size of increments is $ + 1$ . Further,
denoting $H(t)$, $(t > {T_b})$ ,$H(t)$ is the history of the
arrivals up to time $t$ . (Strictly speaking $H( \cdot )$ is a
filtration, an increasing sequence of $\sigma$-algebras.)

\paragraph{Definition 2 (Temporal point process)}A Point process is a collection of mathematical points randomly distributed in
 a mathematical space, such as time and real space. The point process distributed in the time line is called the temporal point process.

A temporal point process is a random process composed of the
sequence of events and the corresponding time stamp ${\rm{\{
}}{t_i}{\rm{\} }}$  these events occurred, where,${{t}_{i}}\in
[{{T}_{b}},{{T}_{e}}]$ , ${T_b}$ is the beginning of the observation
window,${T_e}$ is the end of the observation window. The temporal
point process can be represented as a counting process $N = \{
N(t)|t \in [{T_b},{T_e}]\} $ , where $N(t)$ is a counting process,
records the number of events has happened before time $t$ .Based on
the above definition, multi-dimensional point processes with $C$
types of event can be represented by $C$  counting processes $\{
{N_c}\} _{c = 1}^C$  on a probability space .

A point process can be described via its conditional intensity
function $\left\{ {{\lambda _c}(t)} \right\}_{c = 1}^C$, The
conditional intensity function is also called the hazard function in
survival analysis.

\paragraph{Definition 3 (conditional intensity function)}Conditional intensity functions is $\left\{ {{\lambda_c}(t)} \right\}_{c = 1}^C$, where ${\lambda _c}(t)$ represents the expected instantaneous happening rate that the c-type event occurs instantaneously under a given history record.

\begin{equation}
\begin{array}{l}\label{1}
{{\rm{\lambda }}_c}(t){\rm{ = }}\frac{{{\rm{E(}}N(t + dt) - N(t)|H(t{\rm{))}}}}{{dt}}\\
{\rm{        = }}\frac{{P{\rm{(type\;c\;event\;occurs\;in}}[t,{\rm{ }}t + dt)|H(t{\rm{)}})}}{{dt}}\\
{\rm{        = }}\frac{{P{\rm{(type\;c\;event\;occurs\;in}}[t,{\rm{
}}t + dt)|{\rm{no\;event\;occured\;in
}}[{t_i},t){\rm{,}}H(t{\rm{)}})}}{{dt}}\\
{\rm{        = }}\frac{{P{\rm{(type\;c\;event\;occurs\;in}}[t,{\rm{ }}t + dt),{\rm{no\;event\;in }}[{t_i},t){\rm{|}}H(t{\rm{)}})}}{{P({\rm{no\;event\;occured\;in }}[{t_i},t){\rm{|}}H(t{\rm{)}})dt}}\\
{\rm{        = }}\frac{{p(t,c)}}{{1 - P(t)}}
\end{array}
\end{equation}

Where $p(t,c)$ is the conditional density of type $c$  events at time $t$, $H(t)$ is
the history affects the type $c$ events. ${t_i}$  is the last events'
timestamp before time $t$  and $P(t,{t_i})$ is the conditional cumulative
function, equals the possibility if there were any events happening
in $[{t_i},t)$ . According to \cite{vere2003introduction}, we have

\begin{equation}
\begin{array}{l}\label{2}
p(t,c) = {\lambda _c}(t)\exp ( - \sum\nolimits_{c' = 1}^C {\int_{ti}^t {{\lambda _{c'}}(s)ds} } )\\
P(t,{t_i}) = 1 - \exp ( - \sum\nolimits_{c' = 1}^C {\int_{ti}^t
{{\lambda _{c'}}(s)ds} } )
\end{array}
\end{equation}

For any sequence of events $S = \{ ({t_i},{c_i})\} _{i = 1}^I$, we can calculate its likelihood function as:

\begin{equation}\label{3}
\begin{array}{l}
L(S;\Theta ) = \prod\nolimits_{i = 1}^I {p({t_i},{c_i}) \times (1 - P({T_e},{t_I}))} \\
{\rm{           }} = \prod\nolimits_{i = 1}^I {{\lambda _{{c_i}}}({t_i})}  \times \exp ( - \sum\nolimits_{c = 1}^C {\int_{{T_b}}^{{T_e}} {{\lambda _c}(s)ds} } )
\end{array}
\end{equation}

Different types of point processes have different forms of conditional intensity functions, and we only discuss Hawkes process in this paper.

\subsection{Weibull distribution and exponential distribution in conditional
intensity function}

There is a close relationship between survival analysis and point process analysis. All of these analyses are considering certain event occur at certain time, the most widely used mathematical model is conditional intensity function which is also called hazard function in survival analysis, denote as $h(t)$.

In the survival analysis, survival function is denoted as

\[S(t) = 1 - F(t) = P(T > t)\]
\[F(t) = \int_0^t {f(s)ds} \]

where $F(t)$ is cumulative distribution function,$f(t)$ is probability density function, another important function in survival analysis is hazard function, which is called conditional intensity function in point process analysis, it also represents the expected instantaneous happening rate of some kind of event:

\[h(t) = \mathop {\lim }\limits_{\Delta t \to 0} \frac{{P(t \le T < t + \Delta t|T \ge t)}}{{\Delta t}} = \frac{{f(t)}}{{S(t)}} =  - \frac{d}{{dt}}\log (S(t))\]

Which is corresponding to Eq. (1) and Eq.(2).

In general, the probability distribution commonly used in survival analysis is the Weibull distribution, for example, if we set hazard function $h(t) = \mu \rho {t^{\rho  - 1}}$ , then the corresponding density function $f(t)$ is as follows:

\[f(t) = h(t)\exp ( - \int_0^t {h(s)ds} ) = \mu \rho {t^{\rho  - 1}}\exp ( - \mu {t^\rho })\]

Meanwhile, in the research field of Hawkes process, most of the researchers set these base intensity as a constant, i.e., set ${h_c}(t) = \mu $ , if we only consider influence of an arbitrary event type by its  base intensity, let other type events' base intensity and all the impact function be set as zero, where we denote it as $p(t,c)'$ , which is equivalent to a one-dimensional Poisson process.

The conditional intensity function of Poisson process has following form:

\[\lambda (t) = h(t)\]

note that $h(t)$  is also called hazard function, which could be constant or time-varying function. Then according to (2), we have:

\begin{equation}
  p(t,c)' = {h_c}(t)\exp ( - \int_0^t {{h_c}(s)ds} ) = {\mu _c}\exp ( - {\mu _c}t)
\end{equation}

At this point we can see that, if the base intensity is constant, the c type event influenced by its base intensity is exponentially distributed. However, the exponentially distribution has a critical shortage, i.e., memory-less property, for example:

\begin{equation}
  P(T < s + t|T > t) = \frac{{\int_t^{s + t} {\mu {e^{ - \mu t}}} }}{{\int_t^{ + \infty } {\mu {e^{ - \mu t}}} }} = \frac{{{e^{ - \mu t}} - {e^{ - \mu (t + s)}}}}{{{e^{ - \mu t}}}} = 1 - {e^{ - \mu s}} = P(T < s)
\end{equation}

Characteristics reflected in (5) is corresponding to that the base intensity is time invariant.

However, this assumption is not consistent with the reality. The base possibility that probability of an arbitrary event occurrence should be evolving over time, rather than be a constant. For example, high-quality TV shows will be more popular with evolving over time, and as the kids grow older, young children will become stronger and not get sick easily. If we assume the base intensity is constant, the time invariant model cannot grasp the information mentioned above.

So, to overcome this drawback, and learn the event occurring evolutional tendency with varying time, we introduce Weibull distribution into Hawkes processes, which is widely applied in survival analysis.

 We set $h(t) = \mu \rho {t^{\rho  - 1}}$ , and then we can get corresponding one-dimensional Poisson process $p(t,c)'$:

 \begin{equation}
   p(t,c)' = {h_c}(t)\exp ( - \int_0^t {{h_c}(s)ds} ) = {\mu _c}{\rho _c}{t^{{\rho _c} - 1}}\exp ( - {\mu _c}{t^{{\rho _c}}})
 \end{equation}

 Where ${\mu _c}$ is the scale parameter, which defines the scale of the $h(t)$ , ${\rho _c}$ is the shape parameter that determines whether the function is increasing or decreasing over time. Therefore, under this assumption, the probability of event occurrence influenced by the base intensity obeys the Weibull distribution.

Let's consider a special case, i.e., when we set ${\rho _c}{\rm{ = }}1$ , the base intensity is exponentially distributed. Thus, the exponential distribution is a special case of the Weibull distribution when ${\rho _c}{\rm{ = }}1$, our proposed the Hawkes processes introduced Weibull distribution is more general and take the previous ones as special cases , which is suitable for a broad range of application scenarios.

By observing the parameters obtained after learning the data of the actual event sequence, we can get an estimate of the base instantaneous happening rate at which a certain failure occurs at each time instant. Especially the estimation of shape parameters ${\rho _c}$: if ${\rho _c} < 1$ , then we can know that the base instantaneous happening rate of c kind of failure is descending over time, on the contrary, if ${\rho _c}>1$ , the base instantaneous happening rate of c kind of failure is increasing over time. Then we can obtain the trend of the instantaneous occurrence rate of each type of failure, and the compressors' maintainer and manager can take targeted adjustment measures to reduce the occurrence of failures.

\subsection{Hawkes Processes and Granger Causality}

A C-dimensional point processes generated a set of event sequence, $1, \cdots ,C \in {\cal C}$  are event types. Supposed that there is a subset of event types ${\cal V} \subset {\cal C}$ , for the type $c$ of event, intensity function ${{\rm{\lambda }}_c}(t)$  only influenced by the history of $c$ type of event in ${\cal V}$ ,we denote the history as ${H_{{\cal V}}}(t)$ , the remaining history of event types is denoted by $H_{{\cal C}\backslash {{\cal V}}}(t)
$.In the view of Granger causality,${\cal V}$ amount to the local independence over the dimension of the point process. The occurrence of history events in ${H_{{\cal V}}}(t)$  will affect the probability of occurrence of the future c type of events, and the $H_{{\cal C}\backslash {{\cal V}}}(t)
$ will not. For a subset ${\cal V} \subset {\cal C}$ , set ${N_{\cal V}}{\rm{ = \{ }}{N_c}(t){\rm{|}}c \in {\cal V}{\rm{\} }}$ ,the filtration ${H_{{\cal V}}}(t)$ , i.e., the smallest $\sigma$-algebra ,which is generated by the point process, is defined as $\sigma {\rm{\{ }}{N_c}(s)|s \le t,c \in {\cal V}{\rm{\} }}$. In particular, $H_c(t)$ is the internal filtration of the counting process $N_c(t)$, ${H_{-c}}(t)$ is the filtration for the subset ${\cal C}\backslash \{ c\} $.

\paragraph{Definition 1 \cite{didelez2008graphical}} The counting process $N_c(t)$ is locally independent of $N_{c'}(t)$ given ${N_{C\backslash \{ c,c'\} }}(t)$ if the intensity function ${H_c}(t)$ is measurable with regard to ${H_{-c}}(t)$ for all $t \in [T_b^n,T_e^n]$. Otherwise $N_c(t)$ is locally dependent.

Based on Definition 1, we can apply it to Hawkes processes, and then establish the relationship between the impact function and Granger causality. Definition 1 is equivalent to type $c'$ of event does not have Granger causality with type c of event with regard to $H_(t)$.

Generally speaking, the conditional intensity functions of multi-dimensional Hawkes processes have following form:

\begin{equation}
  \begin{array}{l}
{{\rm{\lambda }}_c}(t) = {h_c}(t) + \sum\nolimits_{c' = 1}^C {\int_0^t {{\phi _{cc'}}(s)d{N_{c'}}(t - s)} } \\
{\rm{       }} = {h_c}(t) + \sum\nolimits_{c' = 1}^C {\int_0^t {{\phi _{cc'}}(t - s)d{N_{c'}}(s)} }
\end{array}
\end{equation}

where ${h_c}(t)$ is the base intensity, which is independent of the history, and in general, ${h_c}(t)$ is a constant function, $\sum\nolimits_{c' = 1}^C {\int_0^t {{\phi _{cc'}}(s)d{N_{c'}}(t - s)} }$ is the endogenous intensity, indicates the history events' effect on the ${\lambda _c}(t)$ . ${\phi _{cc'}}(t)$  is called impact function which measures the influence of historical type $c'$ of events on the type $c$  of events. In this paper, we assume the impact function is stationary, i.e. time-invariant, ${\phi _{cc'}}(t - s) \ge 0,({T_b} \le s < t \le {T_e})$ .Based on this assumption, the work in \cite{eichler2017graphical} reveal the connection between the impact function and Granger causality.

\paragraph{Lemma 1 \cite{eichler2017graphical}}Assume that there exist a Hawkes process with conditional intensity function defined in (7). If the condition $d{N_{c'}}(t - s) > 0,({T_b} \le s < t \le {T_e})$  holds, then if and only if ${\phi _{cc'}}(t) = 0$ for $t \in [0,\infty )$, type $c$ of event and type $c'$ of event don't have Granger causality relationship.
Therefore, for multi-dimensional Hawkes process , if we want to learn its Granger causality between different types of event, we only have to confirm whether the impact function is all zero or not. The Granger causality learning problem is converted to learning impact function problem.

After learning the sequence of failure events, we can get the impact function between various failure events, so that we get the trigger pattern between all failures. In this way, we can take targeted measures to reduce or avoid secondary failure events.

\section{Proposed Hawkes Processes with the Time Varying Weibull base intensity and Learning Algorithm}

In this section, we first introduce the Weibull base intensity to
the Hawkes processes, and then, we change the form of the
conditional intensity function, at last we proposed an efficient
learning algorithm based on MLE method and EM algorithm. Compared
with existing learning algorithms, our algorithm is more effective
in identifying both the base intensity and Granger causality.

\subsection{Introducing the Weibull base intensity to the Hawkes processes}

Following the above analysis and discussion, if we set $h(t) = \mu \rho {t^{\rho  - 1}}$ , then we can get the time varying conditional intensity function of Hawkes process:

\begin{equation}
  \begin{array}{l}
{{\rm{\lambda }}_c}(t){\rm{ = }}{\mu _c}{\rho _c}{t^{{\rho _c} - 1}} + \sum\nolimits_{c' = 1}^C {\int_{{T_b}}^t {{\phi _{cc'}}(t - s)d{N_{c'}}(s)} } \\
{\rm{         }}{\mu _c}{\rho _c}{t^{{\rho _c} - 1}} + \sum\nolimits_{c' = 1}^C {\int_{{T_b}}^t {{\phi _{cc'}}(s)d{N_{c'}}} } (t - s)
\end{array}
\end{equation}

Compared with the normal form of Hawkes process, the base intensity is replaced by ${\mu _c}{\rho _c}{t^{{\rho _c} - 1}}$ with $\mu_c$, then in Hawkes process with time varying conditional intensity function we introduce a new parameters ${\rho _c}$ ,we have to propose a new effective method to learn all the parameters, including ${\rho _c}$.

Here, what we need to emphasize is that the impact function of a certain type of event on itself and its base intensity of the Weibull form are two different concepts. Impact function ${\phi _{cc'}}(t)$  indicates the effect of history events of type c on the occurrence probability of the current type c of event occurrence, and the base intensity of the Weibull form indicates the possibility variation with time for occurrence of the current type c of event. For example, if an elderly person gets sick, then in the future he will become more susceptible to illness because of decreased immunity, and although an elderly person may never get sick, the probability of his getting sick will also increases as he grow older. Note that When we set ${\phi _{cc'}}(t)$ in Eq.(8), this definition in Eq.(8) is equivalent to the normal Hawkes process proposed in the literature, thus the previous proposed point processes model based on Hawkes process is a special case of our proposed ones. Our Hawkes process with the time varying base intensity has better ability to represent the time-varying characteristics of conditional intensity function.

\subsection{Formulating learning Task}

First of all, we parameterize ${\phi _{cc'}}(t) = \sum\nolimits_{m = 1}^M {{a_{cc'm}}{g_m}(t)} $ as described by Lewis, one can refer \cite{lewis2011nonparametric}. Where $g_m(t)$ is the m-th kernel function, which is Gaussian function, and ${a_{cc'}} = ({a_{cc'1}},...,{a_{cc'm}})$ is the corresponding parameters of $g_m(t)$. Suppose that we have a set of event sequence, $S = \{ {s_n}\} _{n = 1}^N$, ${s_n} = \{ (t_i^n,c_i^n)\} _{i = 1}^{{I_n}}$. $t_i^n \in \{ T_b^n,T_e^n\} $is time instant when the i-th event happened in interval $[T_b^n,T_e^n]$ , and $c_i^n \in \{ 1,...,C\} $ ,is the corresponding type of the event. According to the definition of the conditional intensity functions (8) and Eq.(3),the likelihood can be expressed as follows:

\[\begin{array}{l}
{\cal L}({\cal S};\Theta ) = \prod\nolimits_{n = 1}^N {\left\{ {\prod\nolimits_{i = 1}^{{I_n}} {p({t_i},{c_i})}  \times (1 - P(T_e^n))} \right\}} \\
 = \prod\nolimits_{n = 1}^N {\left\{ {\prod\nolimits_{i = 1}^{{I_n}} {{{\rm{\lambda }}_{c_i^n}}{\rm{(}}t_i^n{\rm{)}}}  \times {\rm{exp}}\left( { - \sum\nolimits_{c = 1}^C {\int_{T_b^n}^{T_e^n} {{{\rm{\lambda }}_c}(s)ds} } } \right)} \right\}}
\end{array}\]

Let $\Theta  = \{ {\bf{A}} = [{a_{cc'm}}] \in {\textbf{R}^{C \times C \times M}},{\bf{\mu }} = [{\mu _c}] \in {\textbf{R}^C},{\bf{\rho }} = [{\rho _c}] \in {\textbf{R}^C}\} $,the log-likelihood is:

\begin{equation}
\begin{array}{l}
\log {\cal L}({\cal S};\Theta ) = \sum\nolimits_{n = 1}^N {\left\{ {\sum\nolimits_{i = 1}^{{I_n}} {\log {{\rm{\lambda }}_{c_i^n}}{\rm{(}}t_i^n{\rm{)}} - \sum\nolimits_{c = 1}^C {\int_{T_b^n}^{T_e^n} {{{\rm{\lambda }}_c}{\rm{(}}s{\rm{)}}ds} } } } \right\}} \\
 = \sum\nolimits_{n = 1}^N {\left\{ {\sum\nolimits_{i = 1}^{{I_n}} {\log \left( {{\rho _c}{\mu _c}t_i^{n{\rho _c} - 1}{\rm{ + }}\sum\nolimits_{j = 1}^{i - 1} {\sum\nolimits_{m = 1}^M {{a_{c_i^nc_j^nm}}{g_m}({\tau _{ij}})} } } \right)} } \right.} \\
{\rm{           }}\left. { - \left( {\sum\nolimits_{c = 1}^C {\left( {T{{_e^n}^{{\rho _c}}} - T_b^{n{\rho _c}}} \right){\mu _c}}  + \sum\nolimits_{i = 1}^{{I_n}} {\sum\nolimits_{m = 1}^M {{a_{cc_i^nm}}{G_m}(T_e^n - t_i^n)} } } \right)} \right\}
\end{array}
\end{equation}

Where ${\tau _{ij}} = {t_i} - {t_j}$ ,and ${G_m}(t) = \int_0^t {{g_m}(s)} ds$.

In order to enhance the robustness and accuracy of our proposed model, we consider incorporating the following two regularizers:

\paragraph{Temporal Sparsity} The necessary condition for the stability of the Hawkes process is that $\int_0^\infty  {{\phi _{cc'}}(t) < \infty }$ , which mean that impact function should satisfy the asymptotic stable constraint ${\phi _{cc'}}(t) \to 0$  as $t \to \infty $ \cite{xu2016learning}.Therefore, we add ${L_1}$-norm sparsity regularizer to the parameters of ${g_m}(t)$ , which is denoted as ${\left\| A \right\|_1} = \sum\nolimits_{c,c',m} {\left| {{a_{cc'm}}} \right|}$ .

\paragraph{Local Independence} Based on Lemma 1, if ${\phi _{cc'}}(t) = 0$  for all $t \in [0,\infty )$ , then the type $c'$ event has no effect on the type 'c' event. Thus we incorporate  ${L_{21}}$-norm regularizer \cite{xu2016learning}, \cite{song2013identification, xu2010simple, simon2013sparse}, to constraints the structure of coefficients of ${g_m}(t)$ ,  ${L_{2,1}}$-norm is denoted as ${\left\| A \right\|_{1,2}} = \sum\nolimits_{c,c',m} {{{\left\| {{a_{cc'}}} \right\|}_2}} $ , where ${a_{cc'}} = [{a_{cc'1}},...,{a_{cc'M}}]$ is the corresponding vector of  ${\phi _{cc'}}(t) $. The purpose of incorporating  ${L_{2,1}}$-norm regularizer is to ensure the group sparsity of the coefficient tensor.

Thus, we can get the objective function of the Weibull-Hawkes process is

\begin{equation}
\mathop {\min }\limits_{\Theta  \ge 0} {\rm{ }} - \log {\cal L}({\cal S};\Theta ){\rm{ + }}{\alpha _S}{\left\| A \right\|_1} + {\alpha _G}{\left\| A \right\|_{1,2}}
\end{equation}

\subsection{An EM-based Algorithm}

Similar to \cite{xu2016learning}, \cite{lewis2011nonparametric}, \cite{zhou2013learningb}, \cite{daley2007introduction}, we propose an EM-based learning algorithm to solve the optimization problem (10) iteratively.

\paragraph{Update ${\bf{A}}$ and ${\bf{\mu}}$ } Given the parameters of current step, we can reconstruct the log-likelihood function by Jensen's inequality:

\begin{equation}
\begin{array}{l}
\log ({\mu _{c_i^n}}{\rho _{c_i^n}}{t_i}^{{\rho _{c_i^n}} - 1} + \sum\limits_{j = 1}^{i - 1} {\sum\limits_{m = 1}^M {{a_{c_i^nc_j^nm}}{g_m}(\tau _{ij}^n)} } ) \ge \\
{p_{ii}}\log \frac{{{\mu _{c_i^n}}{\rho _{c_i^n}}{t_i}^{{\rho _{c_i^n}} - 1}}}{{{p_{ii}}}} + \sum\limits_{j = 1}^{i - 1} {\sum\limits_{i = 1}^M {{p_{ijm}}\log (\frac{{{a_{c_i^nc_j^nm}}{g_m}(\tau _{ij}^n)}}{{{p_{ijm}}}})} }
\end{array}
\end{equation}

Where

${p_{ii}} = \frac{{\mu _{c_i^n}^{(k)}\rho _{c_i^n}^{(k)}{t_i}^{\rho _{c_i^n}^{(k)} - 1}}}{{\lambda _{c_i^n}^{(k)}(t_i^n)}}$ and ${p_{ijm}} = \frac{{a_{c_i^nc_j^nm}^{(k)}{g_m}(\tau _{ij}^n)}}{{\lambda _{c_i^n}^{(k)}(t_i^n)}}$

$\lambda _{c_i^n}^{(k)}(t_i^n)$ is the conditional intensity function at $[t_i^n$ with $k$-th step parameters. Then, we can get a tight bound of log-likelihood function as follows:

\begin{equation}
\begin{array}{l}
Q(\Theta {\rm{ }};{\Theta ^{(k)}}) = \sum\limits_{n = 1}^N {\left\{ { - \left( {\sum\limits_{c = 1}^C {\left( {T{{_e^n}^{{\rho _c}}} - T{{_b^n}^{{\rho _c}}}} \right){\mu _c}}  + \sum\limits_{i = 1}^{{I_n}} {\sum\limits_{m = 1}^M {{a_{cc_i^nm}}{G_m}(T_e^n - t_i^n)} } } \right)} \right.} \\
{\rm{                }}\left. { + \sum\limits_{i = 1}^{{I_n}} {\left( {{p_{ii}}\log \frac{{{\mu _{c_i^n}}{\rho _c}{t_i}^{{\rho _c} - 1}}}{{{p_{ii}}}} + \sum\limits_{j = 1}^{i - 1} {\sum\limits_{i = 1}^M {{p_{ijm}}\log \frac{{{a_{c_i^nc_j^nm}}{g_m}(\tau _{ij}^n)}}{{{p_{ijm}}}}} } } \right)} } \right\}
\end{array}
\end{equation}

$\lambda _{c_i^n}^{(k)}(t)$  is the conditional intensity function with current parameters. If and only if $\Theta {\rm{  = }}{\Theta ^{(k)}}$ , we have $Q(\Theta {\rm{ }};{\Theta ^{(k)}}) = \log {\cal L}({\cal S};\Theta )$ .So, we can obtain the surrogate objective function :

\[F =  - Q(\Theta {\rm{ }};{\Theta ^{(k)}}) + {\alpha _S}{\left\| A \right\|_1} + {\alpha _G}{\left\| A \right\|_{1,2}}\]

Then, we can get

\begin{equation}
\begin{array}{l}
\frac{{\partial F}}{{\partial {\mu _c}}} = \sum\limits_{n = 1}^N {\left\{ {\left( {T{{_e^n}^{{\rho _c}}} - T{{_b^n}^{{\rho _c}}}} \right) - \sum\limits_{i = 1}^{{I_n}} {\sum\limits_{{c_i} = c} {{p_{ii}}} } \frac{1}{{{\mu _c}}}} \right\}} \\
\frac{{\partial F}}{{\partial {a_{cc'm}}}} = \sum\limits_{n = 1}^N {\left\{ {\sum\limits_{i = 1}^{{I_n}} {\left( {{G_m}(T_e^n - t_i^n) - \sum\nolimits_{c_i^n = c} {\sum\nolimits_{c_j^n = c'} {\frac{{{p_{ijm}}}}{{{a_{cc'm}}}}} } } \right)} } \right\}} \\
 + {\alpha _S} + {\alpha _G}\frac{{{a_{cc'm}}}}{{{{\left\| {a_{cc'}^{(k)}} \right\|}_2}}}
\end{array}
\end{equation}

Set $\frac{{\partial F}}{{\partial {\mu _c}}} = 0$,and $\frac{{\partial F}}{{\partial {a_{cc'm}}}} = 0$, we can get closed-form solutions of ${\mu _c}$  and ${a_{cc'm}}$:

\begin{equation}
  \begin{array}{l}
{\mu _c}^{(k + 1)} = \frac{{\sum\nolimits_{n = 1}^N {\sum\nolimits_{c_i^n = c} {{p_{ii}}} } }}{{\sum\nolimits_{n = 1}^N {(T{{_e^n}^{{\rho _c}}} - T{{_b^n}^{{\rho _c}}})} }}\\
{a_{cc'm}}^{(k + 1)} = \frac{{ - B + \sqrt {{B^2} - 4AC} }}{{2A}}
\end{array}
\end{equation}

Where

\[\begin{array}{l}
A = \frac{{{\alpha _G}}}{{\left\| {a_{cc'}^{(k)}} \right\|}}\\
B = \sum\nolimits_{n = 1}^N {\sum\nolimits_{c_i^n = c} {{G_m}(T_e^n - t_i^n)} }  + {\alpha _S}\\
C =  - \sum\nolimits_{n = 1}^N {\sum\nolimits_{c_i^n = c} {\sum\nolimits_{c_j^n = c'} {{p_{ijm}}} } }
\end{array}\]

And if ${\alpha _S} = 0$  and ${\alpha _G} = 0$,we have

\begin{equation}
  {a_{cc'm}}^{(k + 1)} = \frac{{\sum\nolimits_{n = 1}^N {\sum\nolimits_{c_i^n = c} {\sum\nolimits_{c_j^n = c'} {{p_{ijm}}} } } }}{{\sum\nolimits_{n = 1}^N {\sum\nolimits_{c_i^n = c'} {{G_m}} (T_e^n - t_i^n)} }}
\end{equation}

\paragraph{Update ${\bm{\rho }}$} Simultaneously,we can get the partial derivative of \emph{F} about ${\bm{\rho }}$:

\begin{equation}
  \begin{array}{l}
\frac{{\partial F}}{{\partial {\rho _c}}} =  - \sum\limits_{n = 1}^N {\left\{ { - {\mu _c}(\ln T_e^n \cdot T{{_e^n}^{{\rho _c}}} - \ln T_b^n \cdot T{{_b^n}^{{\rho _c}}})} \right.} \\
{\rm{          }}\left. { + \sum\limits_{i = 1}^{{I_n}} {\sum\nolimits_{c_i^n = c} {{p_{ii}}(\ln {t_i} + \frac{1}{{{\rho _c}}})} } } \right\}
\end{array}
\end{equation}

If we set $\frac{{\partial F}}{{\partial {\rho _c}}} = 0$ , we cannot get the closed form solution of ${\bm{\rho }}$ . So, we have to update the ${\bm{\rho }}$  by the gradient descent method:

\begin{equation}
  {\rho _{k + 1}} = {\rho _k} - {\alpha _\rho }\frac{{\partial F}}{{\partial {\rho _c}}}
\end{equation}

Where ${\alpha _{\bm{\rho }}}$ is the learning rate of ${\bm{\rho }}$ , and in order to determine when the iterative processes terminates, we adopt a strategy called early stopping \cite{caruana2001overfitting}, i.e., if the log-likelihood on the validation data with current parameters will decent, then the iterative processes will be terminated.
Here is the algorithm of estimating the parameters of the Weibull-Hawkes process.

\begin{algorithm}
    \renewcommand{\algorithmicrequire}{\textbf{Input:}}
    \renewcommand{\algorithmicensure}{\textbf{Output:}}
    \caption{EM-based Algorithm for Weibull-Hawkes process (WBMLE-SGL))}
    \label{alg:1}
    \begin{algorithmic}[1]
        \REQUIRE Event sequences $\{ S_n \} _{n = 1}^N$, trade-off parameters $\alpha_S$,$\alpha_G$,${\alpha _{\bm{\rho }}}$ and $k$.
        \ENSURE Parameters of model $\bm{\mu}$,${\bm{\rho }}$ and
$\textbf{\emph{A}}$.
        \STATE  Initialize $ \emph{\textbf{A}} = [a_{cc'm} ] $ and $\bm{\mu}=[\mu_c]$  randomly,Set ${\bm{\rho }} = [{\rho _c}],{\rho _c} = 1$ .
        \REPEAT
        \REPEAT
        \STATE Update $\bm{\mu}$ and $\emph{\textbf{A}}$ via Eq. (14) and Eq. (15) respectively.
        \UNTIL{Convergence}
        \STATE Update ${\bm{\rho }}$  via Eq. (17) $k$  times.
        \UNTIL{The likelihood function is not changing or satisfy the early stopping condition.}
    \end{algorithmic}
\end{algorithm}

\section{Experimental results}

To verify the robustness and effectiveness of our proposed Hawkes processes with the time varying base intensity more accurately, we compare our model with the state-of-the-art model i.e., MLE-SGLP \cite{xu2016learning}, which can reveal the Granger causality robustly and accurately on synthetic datasets and real-world datasets with constant base intensity. First of all, we generate sets of event sequence both with time-varying base intensity and constant base intensity, respectively. Then we test our proposed Weibull-Hawkes process model and compare it to the MLE-SGLP and MLE model on these two kinds of synthetic datasets. In order to figure out the influence of regularizer, we consider simultaneously two learning scenarios: the pure Weibull-Hawkes process without any regularizer (WB), the Weibull-Hawkes process with spare regularizer and group-lasso regularizer (WB-SGL).
To make the evaluation of different models more intuitive, we apply the following measure criteria:

  1) The log-likelihood of testing data, denote as Loglike;

  2) The relative error of ${{\bm{\mu}}}$:\[{e_{{\bm{\mu}}}} = \frac{{{{\left\| {{\bf{\tilde \mu }} - {{\bm{\mu}}}} \right\|}_2}}}{{{{\left\| {{\bm{\mu}}} \right\|}_2}}}\]

  3) The relative error of ${{\bm{\rho}}}$:\[{e_{{\bm{\mu}}}} = \frac{{{{\left\| {{\bf{\tilde \rho }} - {{\bm{\rho}}}} \right\|}_2}}}{{{{\left\| {{\bm{\rho}}} \right\|}_2}}}\]

  4)In order to testify the error of the base intensity function, we denote a relative error of $h(t)$ as:\[{e_{h(t)}} = \frac{1}{C}\sum\nolimits_{c \in C} {\frac{{\int_{{T_b}}^{{T_e}} {\left| {{{\tilde h}_c}(t) - {h_c}(t)} \right|dt} }}{{\int_{{T_b}}^{{T_e}} {{h_c}(t)} dt}}} \]

  5) Because there are some impact functions which are all-zero, the relative error of impact functions do not exist, instead, we choose the absolute error of impact functions:
      \[\Phi (t) = [{\phi _{cc'}}(t)]\]
      \[{e_\Phi } = \sum\nolimits_{c,c'} {\int_0^T {\left| {{{\tilde \phi }_{cc'}}(t) - {\phi _{cc'}}(t)} \right|} } dt\]

  6)  Accuracy of granger causality analysis via distinguishing the impact function which is all-zero or not.

After testing a variety of patterns with the synthetic data experiment, we test our proposed models on real-world data: the failure event sequence of 87 compressor station during twelve years. Learning our model on these data reveals a lot of useful information in strategizing the actual operation and maintenance of the compressor station. We will introduce it in section 5.4.

\subsection{Data generating protocol}

To assess the usefulness of our model, we generate the time-varying base intensity data \cite{ogata1981lewis}. Our data has the two kinds of impact function: sine-like impact functions and square-like impact functions. Each of them is a 5-dimensional Hawkes process, and contains 500 asynchronous event sequences with time length 50. The ${\bm{\mu }}$  of each event type is uniformly sampled from $[0,\frac{1}{5}]$, the ${\bm{\rho}}$ of each event type is uniformly sampled from  \[[0.5,1.5]\].The sine-like impact functions are generated as

\[\phi cc'(t) = \left\{ \begin{array}{l}
{A_{cc'}}(1 - \cos ({\omega _{cc'}}t + {\varphi _{cc'}})),\,{\rm{    }}t \in [0,\frac{{2\pi  - {\varphi _{cc'}}}}{{{\omega _{cc'}}}}]\\
0,{\rm{                                         otherwise}}
\end{array} \right.{\rm{    }}c,c' \in 1,...,5\]

Where If $c' \in {\rm{\{ 1,2,3\} }}$ and $c \in {\rm{\{ 1,2,3\} }}$ , then ${A_{cc'}} = 0.05,{\omega _{cc'}} = 0.6\pi ,{\varphi _{cc'}} = 0$ ; If $c' \in {\rm{\{ 4,5\} }}$ and $c \in {\rm{\{ 4,5\} }}$ , then ${A_{cc'}} = 0.05,{\omega _{cc'}} = 0.4\pi ,{\varphi _{cc'}} = \pi $; If $c' \in {\rm{\{ 5\} }}$ and $c \in {\rm{\{ 1,2,3\} }}$ or $c \in {\rm{\{ 5\} }}$  and $c' \in {\rm{\{ 1,2,3\} }}$ , then ${A_{cc'}} = 0$ ; If $c' \in {\rm{\{ 4\} }}$ and $c \in {\rm{\{ 1,2,3\} }}$ or $c \in {\rm{\{ 4\} }}$ and $c' \in {\rm{\{ 1,2,3\} }}$ , then ${A_{cc'}} = 0.02,{\omega _{cc'}} = 0.2\pi ,{\varphi _{cc'}} = \pi $. The square-like impact functions are the truncated results of above sine-like impact functions. In order to testify the MLE-SGLP model, we set three pairs of different events type: [1, 2; 2, 3; 4, 5], which pairs of events type have the similar Granger causality.

\begin{algorithm}
    \renewcommand{\algorithmicrequire}{\textbf{Input:}}
    \renewcommand{\algorithmicensure}{\textbf{Output:}}
    \caption{Generating Sequence Data of Hawkes Processes}
    \label{alg:2}
    \begin{algorithmic}[2]
        \REQUIRE scale parameter ${{\bm{\mu}}}$, shape parameter ${{\bm{\rho}}}$, impact funcitons $\Phi (t) = [{\phi _{cc'}}(t)]$ and length of sequence $T$.
        \ENSURE Event sequence $S_n$.
        \STATE  Initialize $ {\rm{t}} = \varepsilon {\rm{ }}(\varepsilon {\rm{  > 0}})$, $\varepsilon$ is a small positive number, ${S_n} = \emptyset$.
        \REPEAT
        \STATE Calculate $M = {\max } \sum\nolimits_1^C {{\lambda _{{\rm{ }}c}}(t')} $ ,
        \STATE where $t' \in [t,t + \max ({T_{cc'}})]$, $T_{cc'}$ is the max period of all the ${\phi _{cc'}}(t)$,
        \STATE and ${\lambda _{{\rm{ }}c}}(t'){\rm{ = }}{\mu _c}{\rho _c}t{'^{{\rho _c} - 1}} + \sum\limits_{j = 1}^{i - 1} {\sum\limits_{m = 1}^M {{a_{c_i^nc_j^nm}}{g_m}(t' - t)} } $
        \STATE Generate an exponential r.v. $E$ with mean $\frac{1}{M}$ and an r.v. $U$ uniformly distributed on (0, 1).
        \REPEAT
        \STATE Calculate $o=\sum\limits_{c = 1}^{c'} {{\lambda _c}(t + E)}$,$c' \in C$.
        \UNTIL{$o\ge M \cdot U$}
        \STATE $t \leftarrow t + E$,${S_n} = {S_n} \cap \{ (t,c')\} $
        \UNTIL{$t > T$}
    \end{algorithmic}
\end{algorithm}

Based on the Algorithm 2, we generate a set of event sequences we have mentioned. figure 2 depicts conditional intensity function and event-occurrence time of 5 types' event sequences.

\begin{figure}[htbp]
\centering \subfigure[sine-impact funciton]{
\begin{minipage}[t]{1.0\linewidth}
\centering
\includegraphics[width=1.0\linewidth]{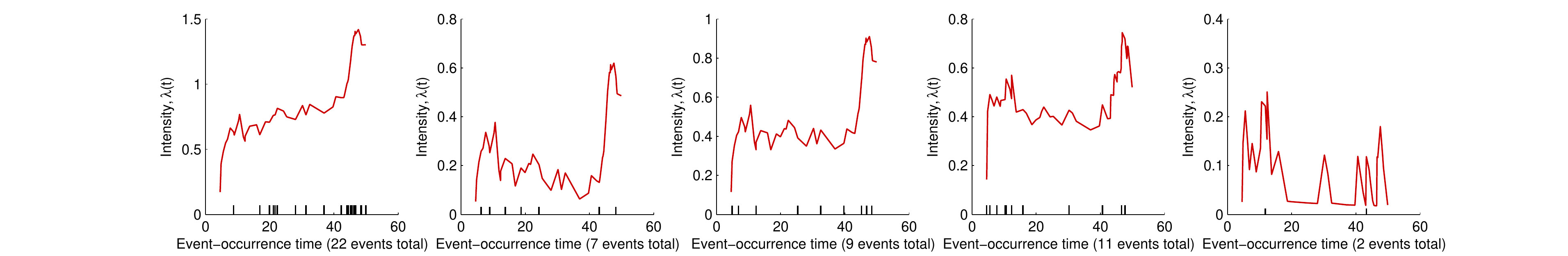}
\end{minipage}%
}%

\quad

\subfigure[square-impact funciton]{
\begin{minipage}[t]{1.0\linewidth}
\centering
\includegraphics[width=1.0\linewidth]{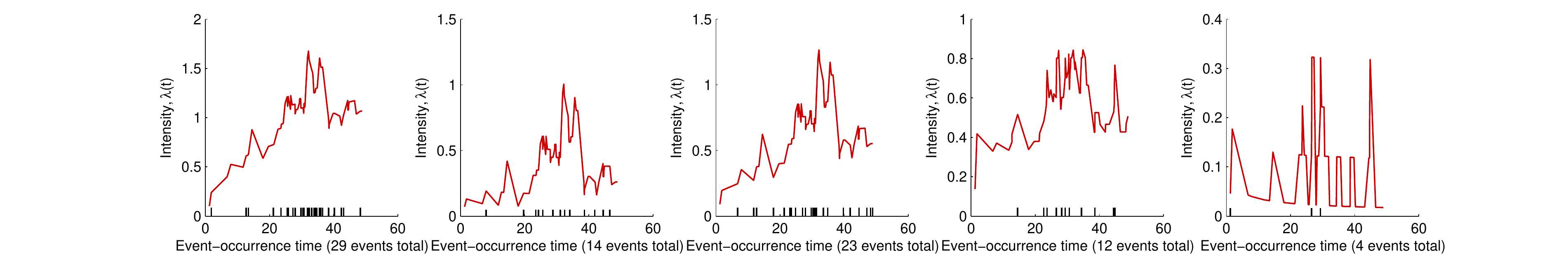}
\end{minipage}%
}%

\centering \caption{The conditionnal intensity function and
event-occurrence time. The red curve above represents the intensity
function, black points represent the time stamp of the current
event.}
\end{figure}

figure 2 discribe the condition intensity functions' curves and when and which type the event occurred. We can find-out that the higher conditional intensity function, the higher possibility of the event occurred synchronously. The following test of our model will be executed on these event sequences.

\subsection{Experimental results of time-varying base intensity data}
\subsubsection{The choice of hyper-parameters}

First of all, we are supposed to test our model with various parameters in a wide range,${\alpha _S},{\alpha _G} \in [{10^{ - 2}},{10^3}]$, the curves of Loglike with regard to the two parameters are shown in following four subfigure in figure 3, the upper part is sine-impact function and the lower part is square-impact function.

\begin{figure}[htbp]
\centering

\subfigure[Sine-impact function and ${\alpha_S}$]{
\begin{minipage}[t]{0.5\linewidth}
\centering
\includegraphics[width=2.5in]{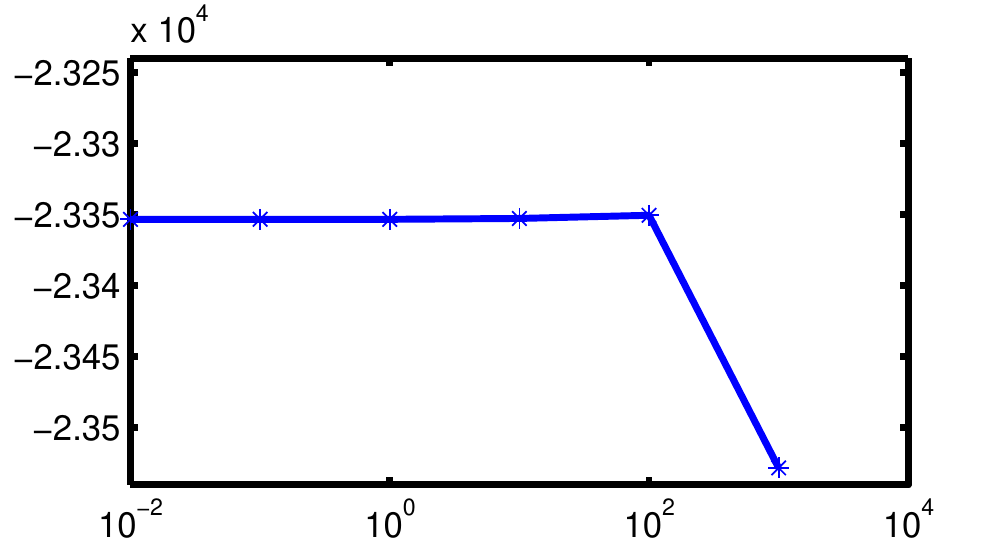}
\end{minipage}%
}%
\subfigure[Sine-impact function and ${\alpha_G}$]{
\begin{minipage}[t]{0.5\linewidth}
\centering
\includegraphics[width=2.5in]{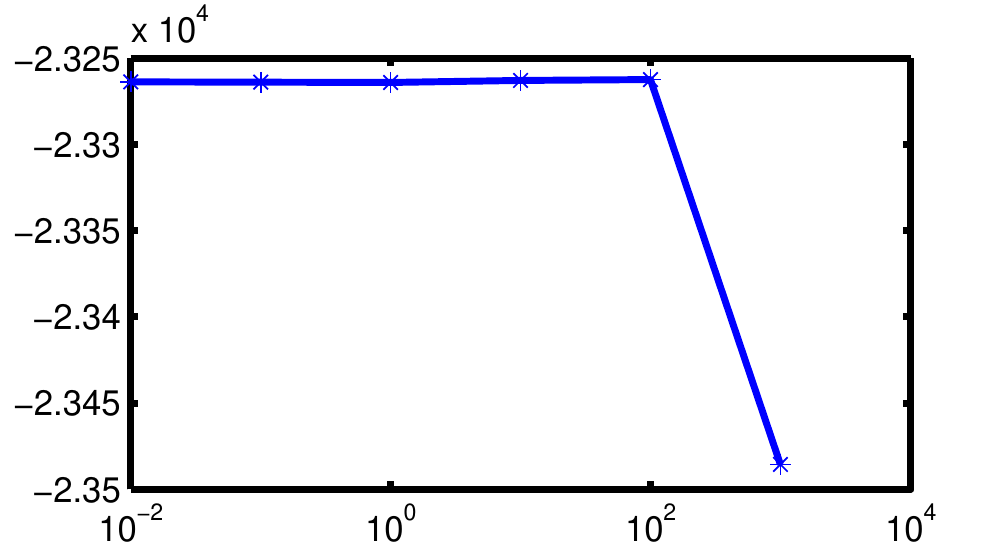}
\end{minipage}%
}%
\quad \subfigure[Square-impact function and ${\alpha_S}$]{
\begin{minipage}[t]{0.5\linewidth}
\centering
\includegraphics[width=2.5in]{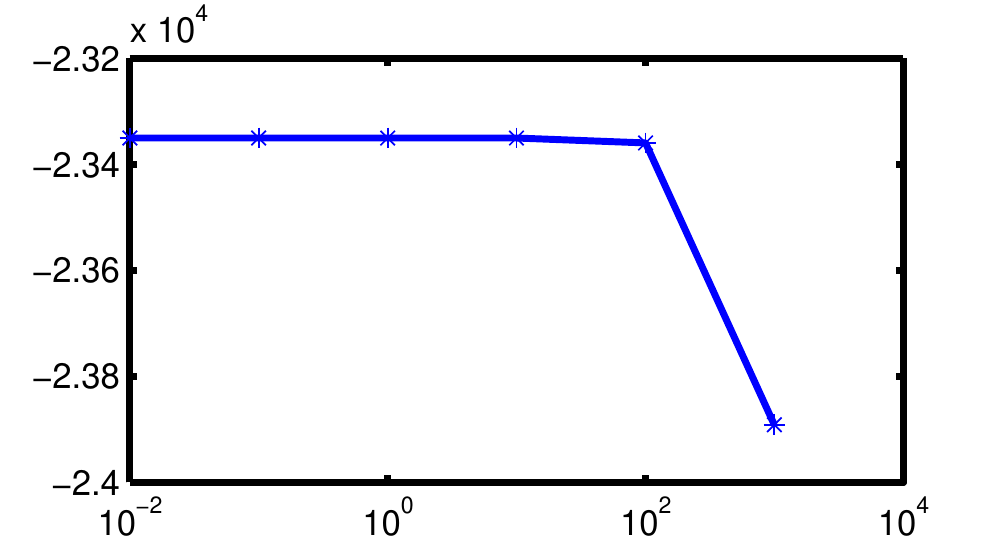}
\end{minipage}
}%
\subfigure[Square-impact function and ${\alpha_G}$]{
\begin{minipage}[t]{0.5\linewidth}
\centering
\includegraphics[width=2.5in]{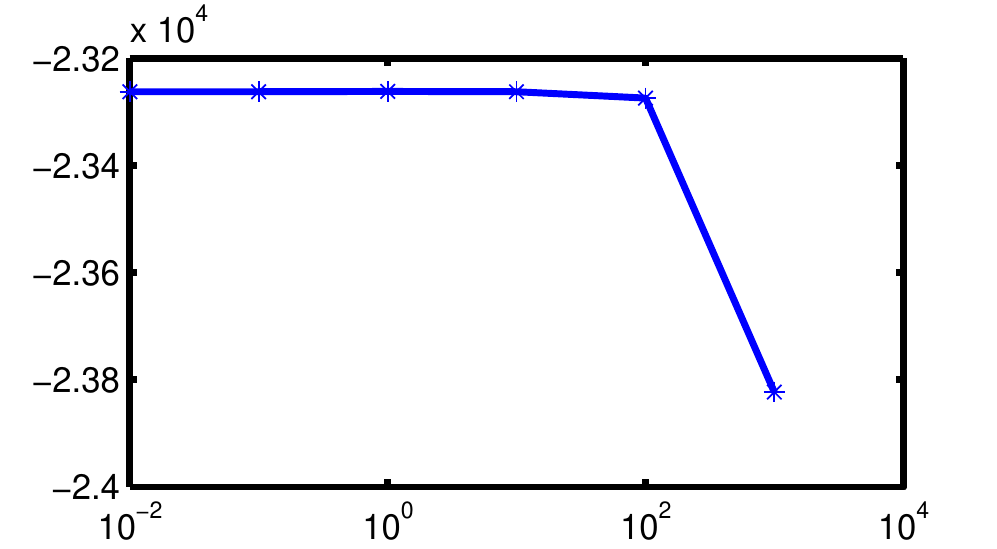}
\end{minipage}
}%

\centering \caption{The curves of Loglike with the change of
${\alpha_S}$ and ${\alpha_G}$}
\end{figure}

When ${\alpha _{\rm{G}}}$ is changing within the range of $[{10^{ - 2}},{10^3}]$ ,  ${\alpha _{\rm{S}}}$ is fixed at 10, when ${\alpha _{\rm{S}}}$ is changing within the range of $[{10^{ - 2}},{10^3}]$, ${\alpha _{\rm{G}}}$  is fixed at 100.We can find out that our model is relatively stable when the hyper-parameters is changing in a wide range. According to the experimental result, we can set ${\alpha _{\rm{S}}} = 10$ , and ${\alpha _{\rm{G}}} = 100$.

\subsubsection{Relative error of parameters
}

In the following subsections, we will compare the performance of different models on all kinds of measure criteria, such as ${e_{\bf{\mu }}}$,${e_{\bf{\rho }}}$,${e_{h(t)}}$,${e_\Phi }$and Loglike.

\begin{figure}[htbp]
\centering
\subfigure[Sine-impact function]{
\begin{minipage}[t]{0.5\linewidth}
\centering
\includegraphics[width=0.9\linewidth]{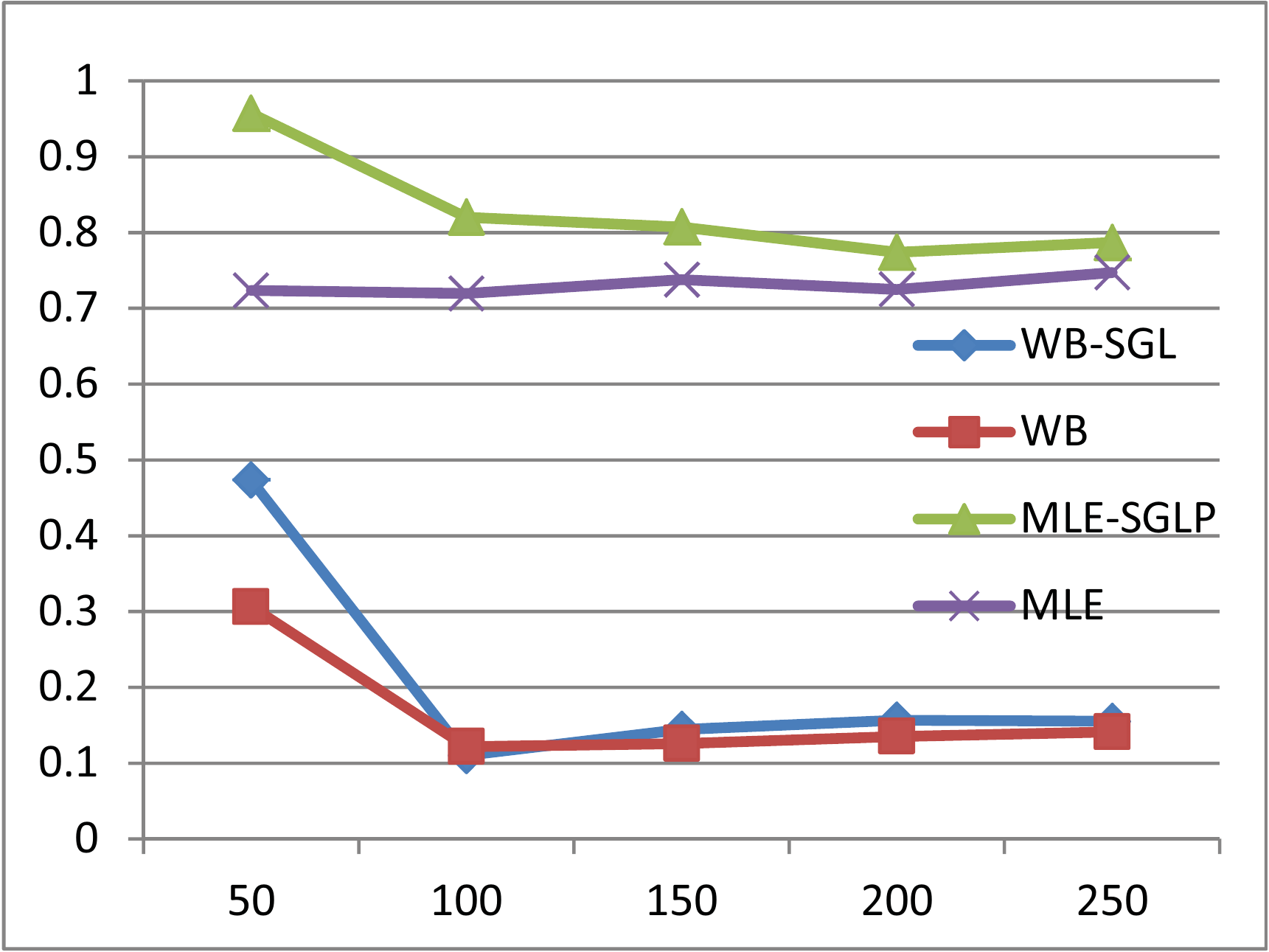}
\end{minipage}%
}%
\subfigure[Square-impact function]{
\begin{minipage}[t]{0.5\linewidth}
\centering
\includegraphics[width=0.9\linewidth]{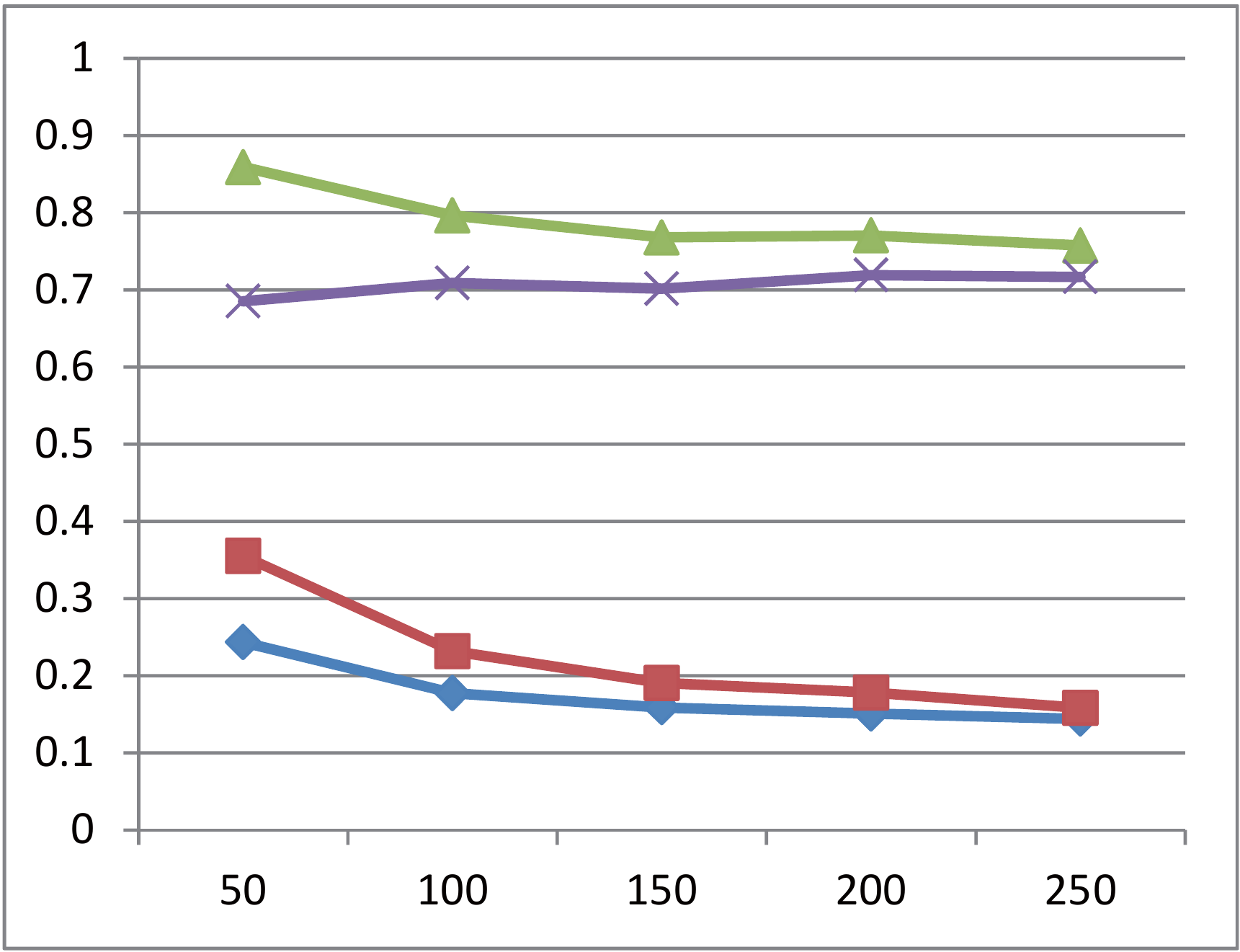}
\end{minipage}%
}%
\centering \caption{The curves of ${e_{\bf{\mu }}}$ of the four sorts of models: WB,WB-SGL, MLE, and MLE-SGLP.}
\end{figure}

Based on discussion in subsection 4.1, we know that MLE models fix the value of $\rho_c$ to 1, so these models cannot fit the $h_c(t)$ well, intuitionally, we anticipate that the relative error of scale parameter $\bm{\mu}$ for MLE will bigger than our WB model. The estimate of $\bm{\mu}$ for MLE model is misled by setting the $\rho_c$ as fixed value, In figure 4, it is shown that the estimated relative error ${e_{\bf{\mu }}}$ of the parameter $\bm{\mu}$ with our model is much smaller than the MLE model. With increasing the number of training sequences, the performance of our model has improved greatly, however the pure MLE model have an increasing relative error, because the time-varying base intensity interferes the estimation of the impact functions and $\bm{\mu}$ without the regularizers.

\begin{figure}[htbp]
\centering
\subfigure[Sine-impact function]{
\begin{minipage}[t]{0.5\linewidth}
\centering
\includegraphics[width=0.9\linewidth]{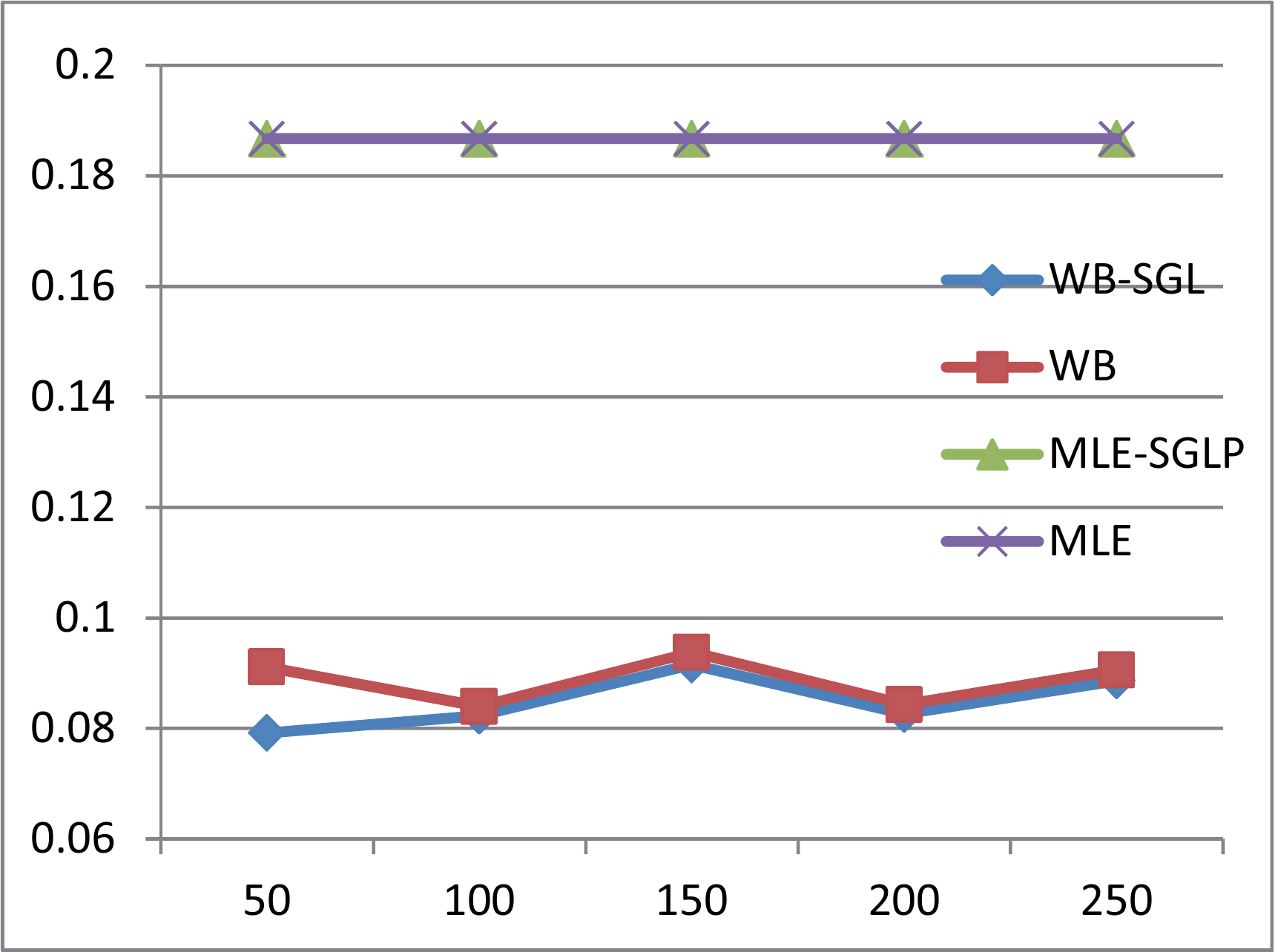}
\end{minipage}%
}%
\subfigure[Square-impact function]{
\begin{minipage}[t]{0.5\linewidth}
\centering
\includegraphics[width=0.9\linewidth]{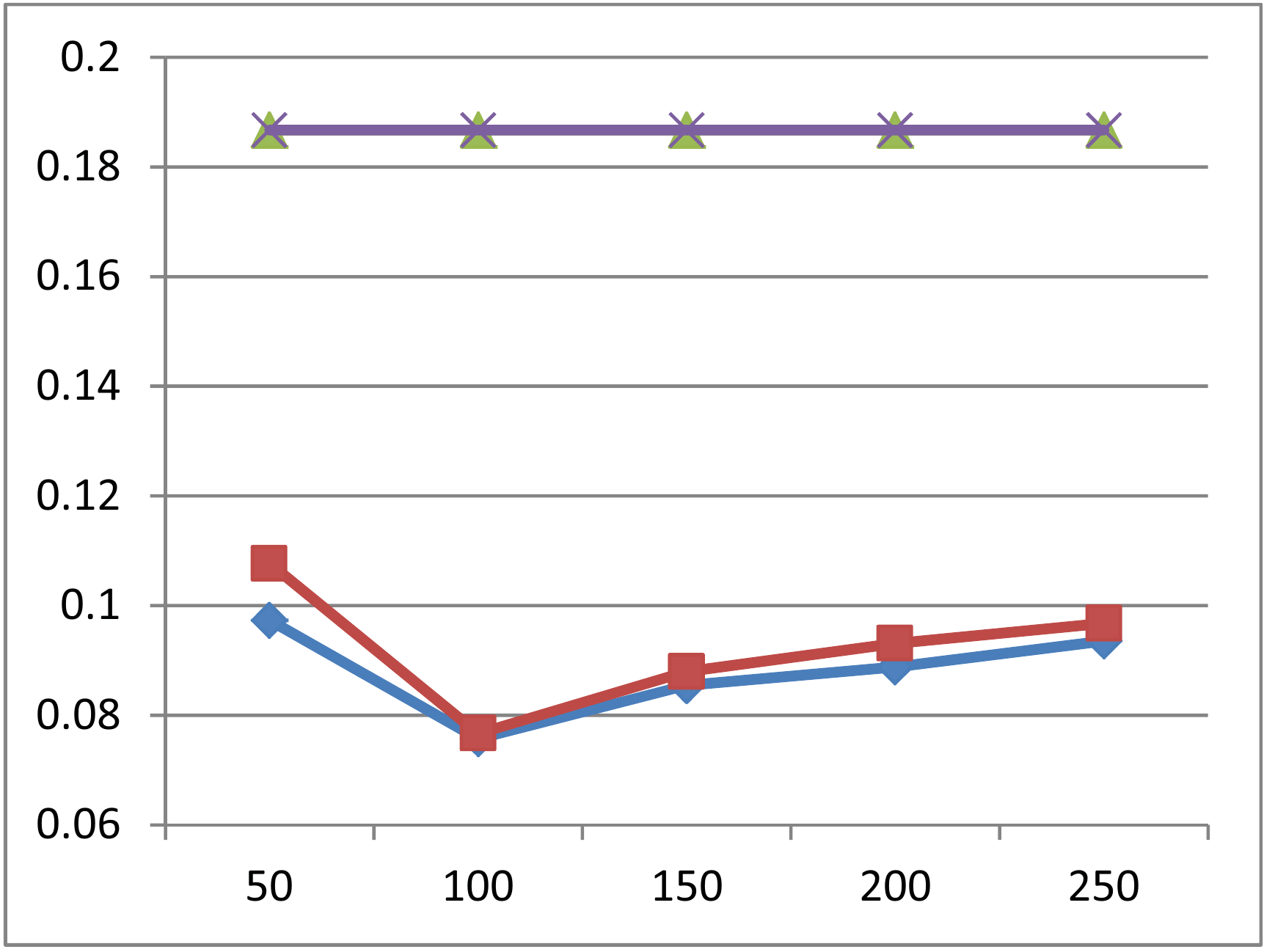}
\end{minipage}%
}%
\centering \caption{The curves of ${e_{\bf{\rho }}}$ of the four sorts of models: WB,WB-SGL, MLE, and MLE-SGLP.}
\end{figure}

As mentioned above, all of the MLE-based algorithms assume the scale parameter is fixed to $\rho_c=1$ , so the relative error of $\bm{\rho}$ is fixed to 0.18674, while our WB-based models can accurately learn parameter $\bm{\rho}$ from the asynchronous event sequences (see Figure 6). The fluctuation of $\bm{\rho}$ is due to the randomness of the event sequences. It is still difficult to get an unbiased estimate of $\bm{\rho}$.

\begin{figure}[htbp]
\centering
\subfigure[Sine-impact function]{
\begin{minipage}[t]{0.5\linewidth}
\centering
\includegraphics[width=0.9\linewidth]{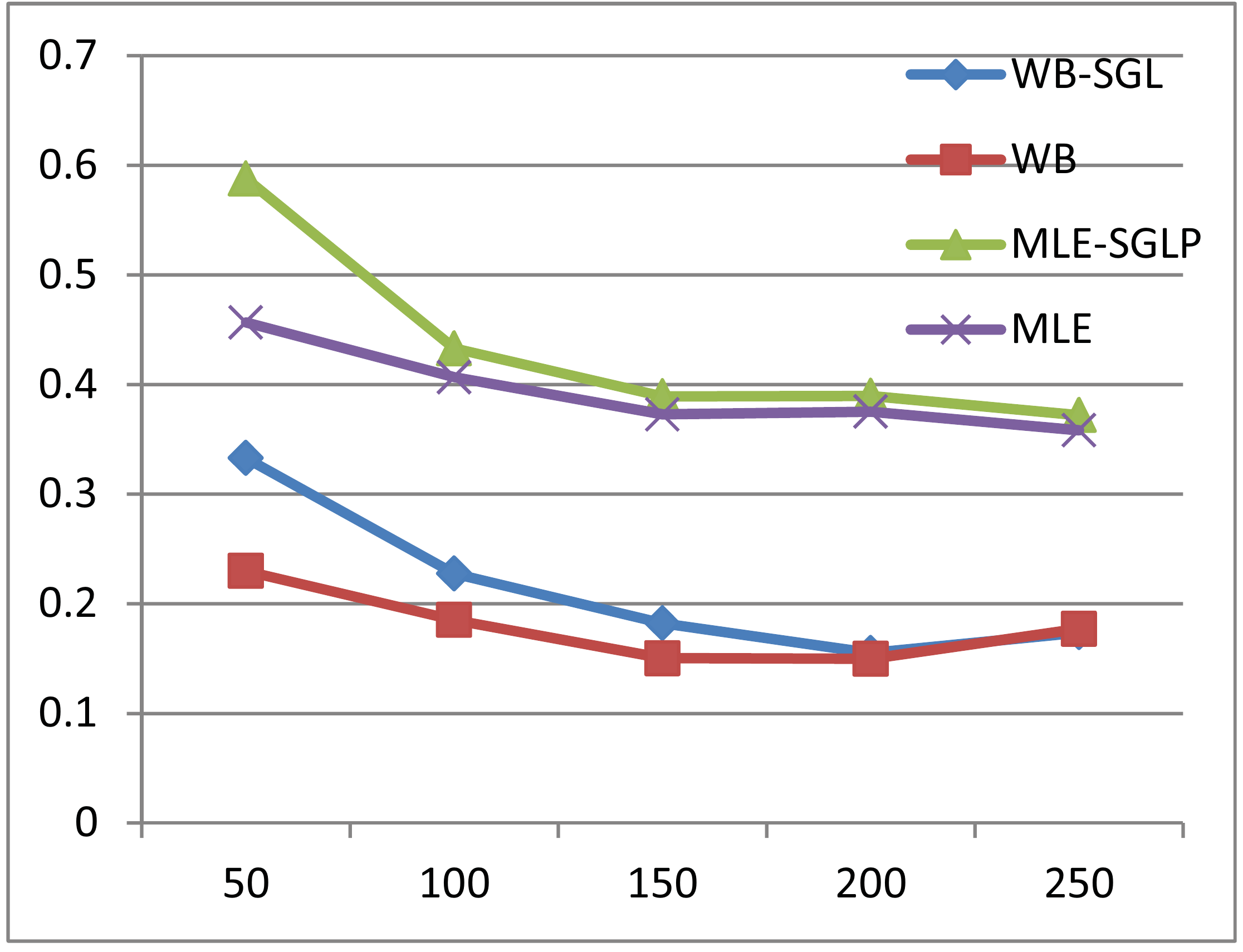}
\end{minipage}%
}%
\subfigure[Square-impact function]{
\begin{minipage}[t]{0.5\linewidth}
\centering
\includegraphics[width=0.9\linewidth]{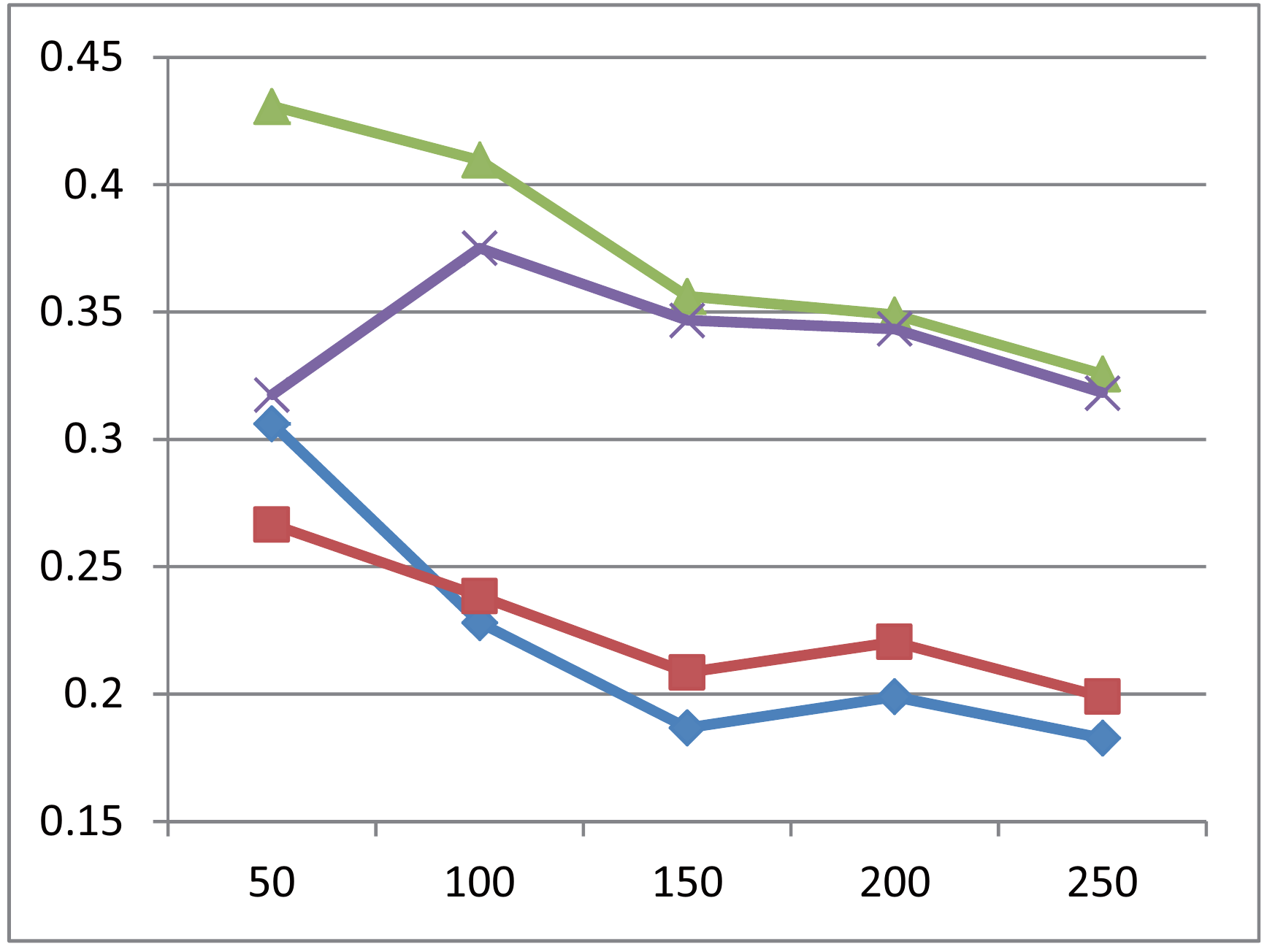}
\end{minipage}%
}%
\centering \caption{The curves of $e_{h(t)}$ of the four sorts of models: WB,WB-SGL, MLE, and MLE-SGLP.}
\end{figure}

Comparing the scale parameter and the shape parameter of different algorithms separately might be relatively unfair for the MLE-based model (because in fact there are only a base intensity parameter which is needed to be estimated), thus, we first define a new measure criterion,the relative error of the base intensity ${e_{h{(t)}}}$, by comparing the estimate accuracy of the base intensity , we can figure out which approach's estimate for $h(t)$ is better. From the Figure 7 we can find that, with increasing the number of training sequences, the relative error of all the algorithms is descending, but limited by model assumptions, MLE-Based model cannot fit the base intensity $h{(t)}$ well. The estimates of $h{(t)}$ with our models are superior to the MLE models.

\begin{figure}[htbp]
\centering
\subfigure[Sine-impact function]{
\begin{minipage}[t]{0.5\linewidth}
\centering
\includegraphics[width=0.9\linewidth]{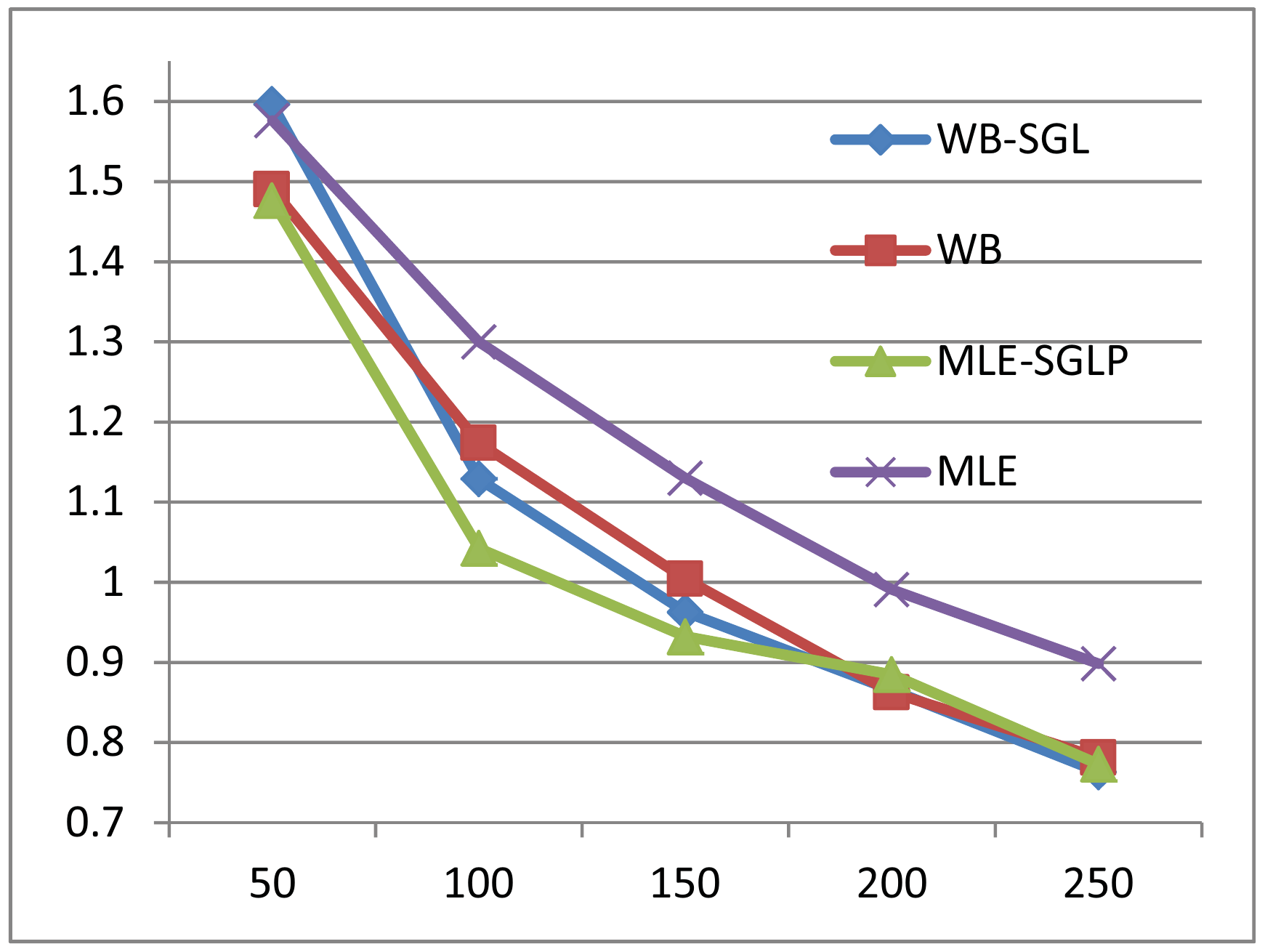}
\end{minipage}%
}%
\subfigure[Square-impact function]{
\begin{minipage}[t]{0.5\linewidth}
\centering
\includegraphics[width=0.9\linewidth]{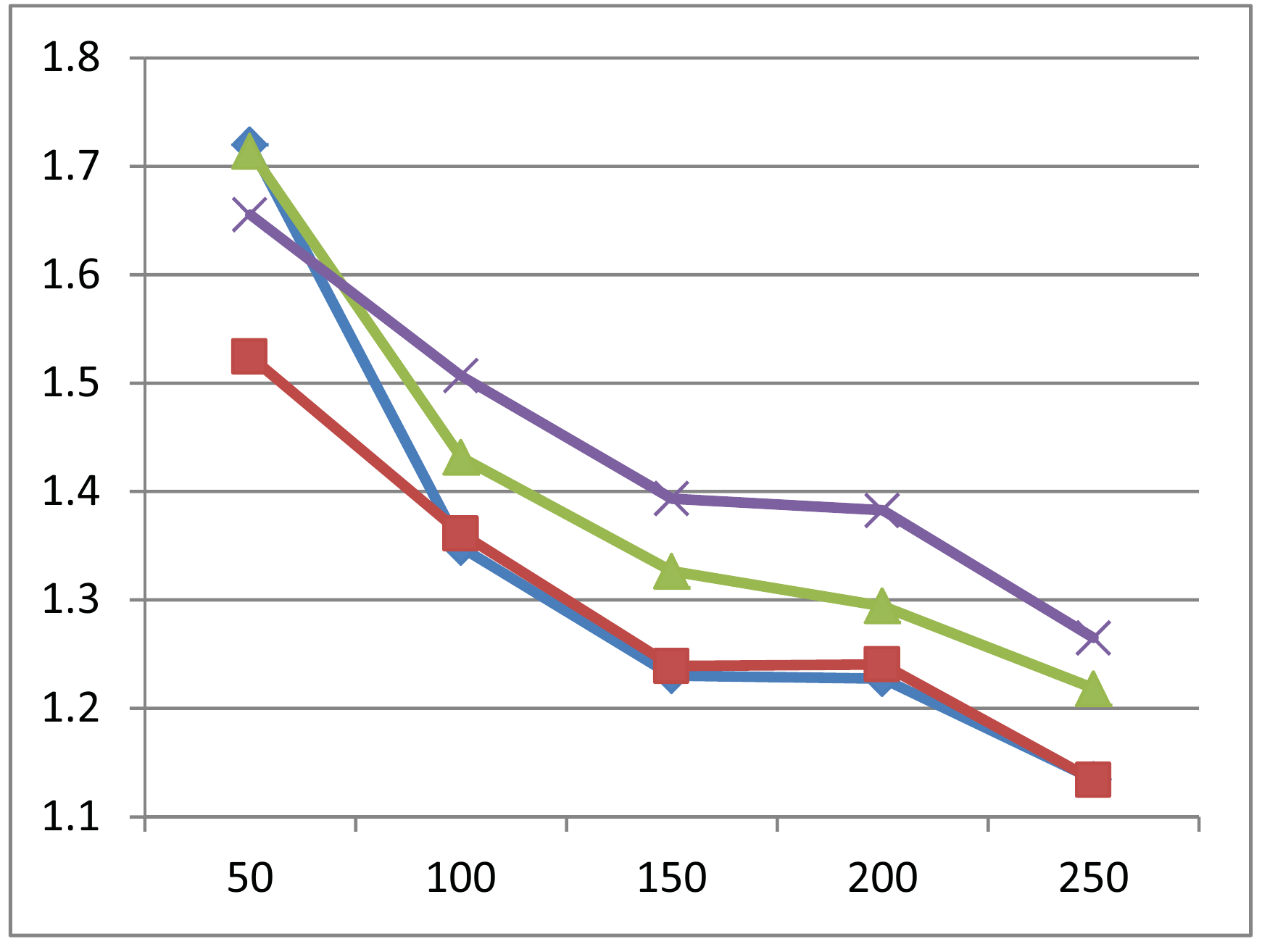}
\end{minipage}%
}%
\centering \caption{The curves of $e_{\Phi}$ of the four sorts of models: WB,WB-SGL, MLE, and MLE-SGLP.}
\end{figure}

In this subsection, we will compare the estimation error of impact functions with the four sorts of models: WB,WB-SGL, MLE, and MLE-SGLP, which is the most important part to reflect Granger causality relationship between events. From Figure 7, we can find that, in sine-like case, when the number of training sequences is 250, the estimation error values of $e_{\Phi}$  with WB-SGL, WB and MLE-SGLP algorithm are 0.763295, 0.781096 and 0.7729, respectively. Meanwhile, in square-like case, we can figure out that the estimate of the impact function with WB-SGL algorithm is the most accurate, however, due to the influence of high frequency components in square wave, the performance of the method based on Gauss basis function is affected. The $e_{\Phi}$  values of all MLE-based algorithms are affected by the larger estimating errors of the base intensity $h(t)$ .

\begin{figure}[htbp]
\centering
\subfigure[Sine-impact function]{
\begin{minipage}[t]{0.5\linewidth}
\centering
\includegraphics[width=0.9\linewidth]{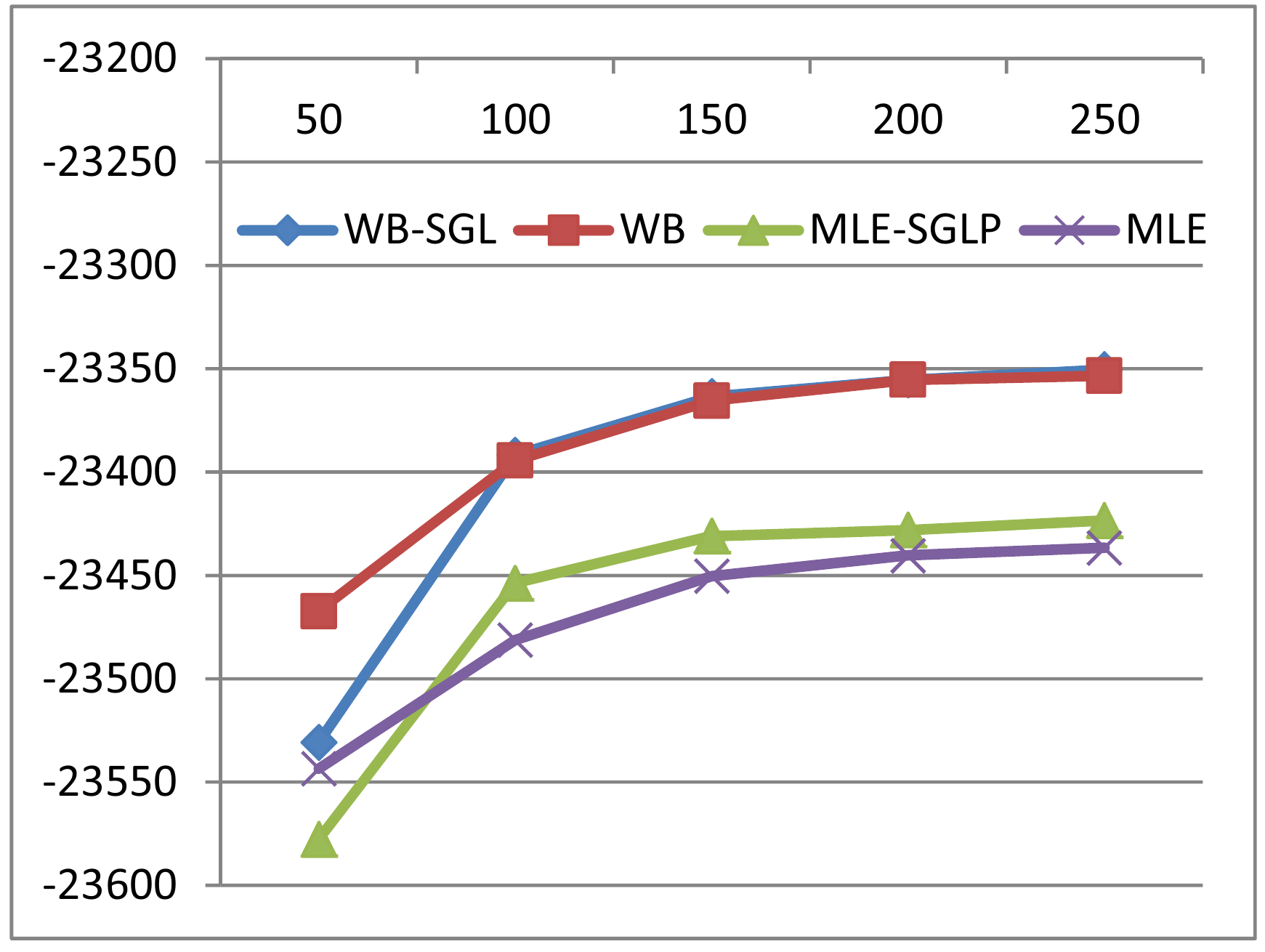}
\end{minipage}%
}%
\subfigure[Square-impact function]{
\begin{minipage}[t]{0.5\linewidth}
\centering
\includegraphics[width=0.9\linewidth]{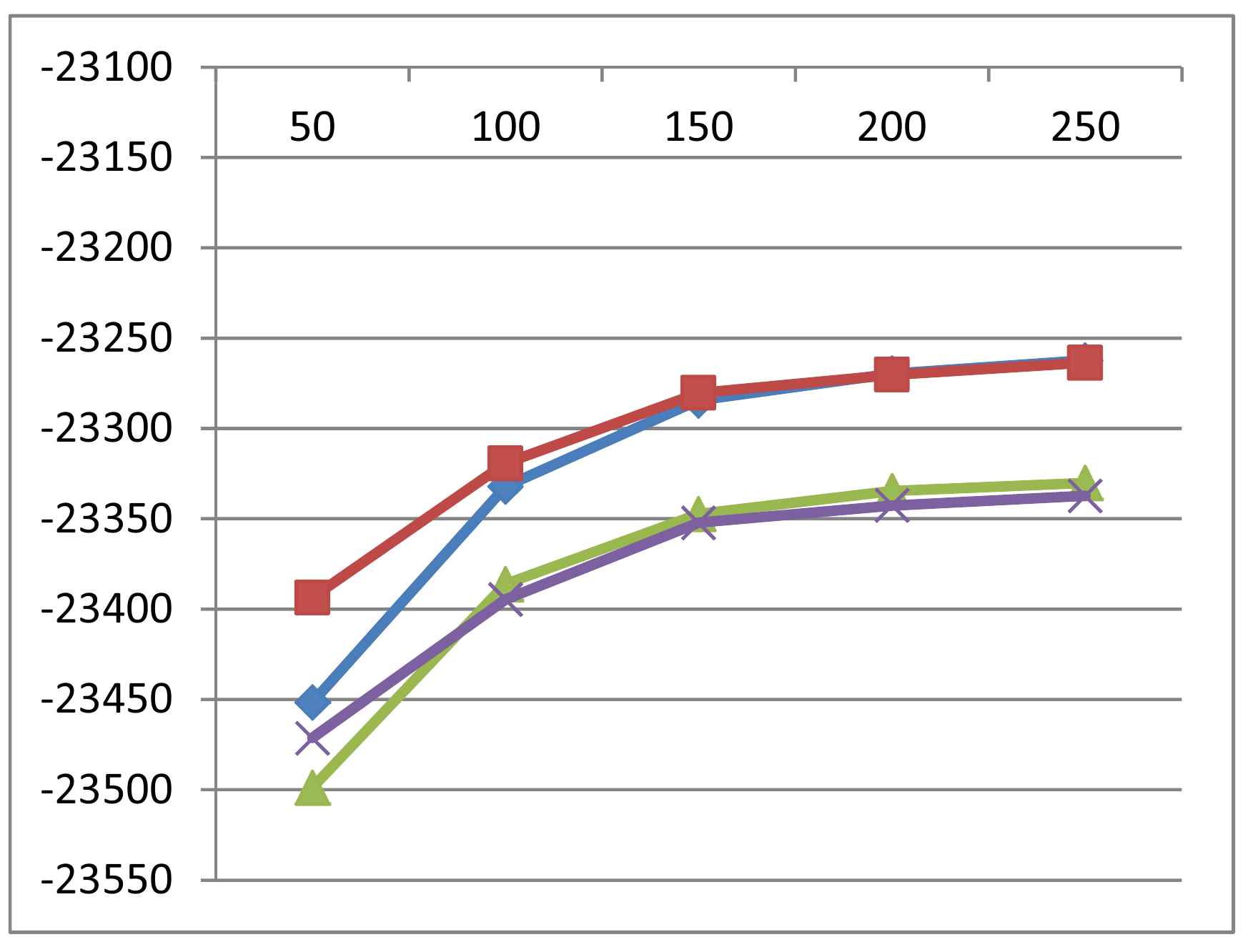}
\end{minipage}%
}%
\centering \caption{The curves of Loglike of the four sorts of models: WB,WB-SGL, MLE, and MLE-SGLP.}
\end{figure}

Based on the above analysis, we can validate that when base intensity $h(t)$ is time-varying, WB-based models are superior to the MLE-based ones. The curves of Loglike also verify that. From Figure 8, we can see that the Loglike of WB-SGL and WB algorithm are much larger than MLE-SGLP and MLE algorithm. Specifically, WB-SGL is better than WB algorithm, both in sine-like and square-like case.

\subsubsection{Curves of the impact functions
}
\begin{figure}[htbp]
\centering \subfigure[Sine-impact function]{
\begin{minipage}[t]{1.0\linewidth}
\centering
\includegraphics[trim = 40mm 0mm 0mm 0mm,width=1.0\textwidth]{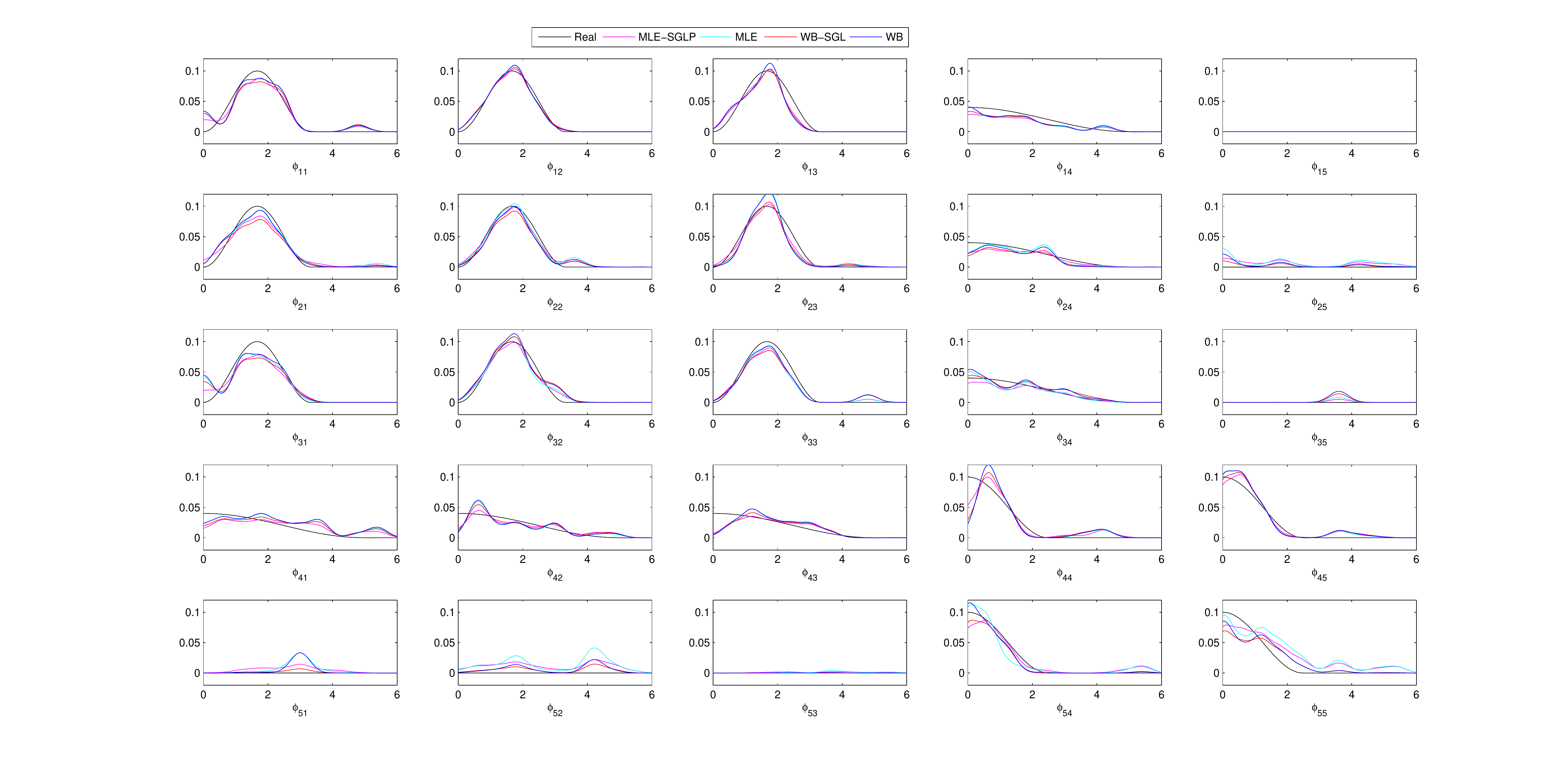}
\end{minipage}%
}%
\quad \subfigure[Square-impact function]{
\begin{minipage}[t]{1.0\linewidth}
\centering
\includegraphics[trim = 40mm 0mm 0mm 0mm,width=1.0\textwidth]{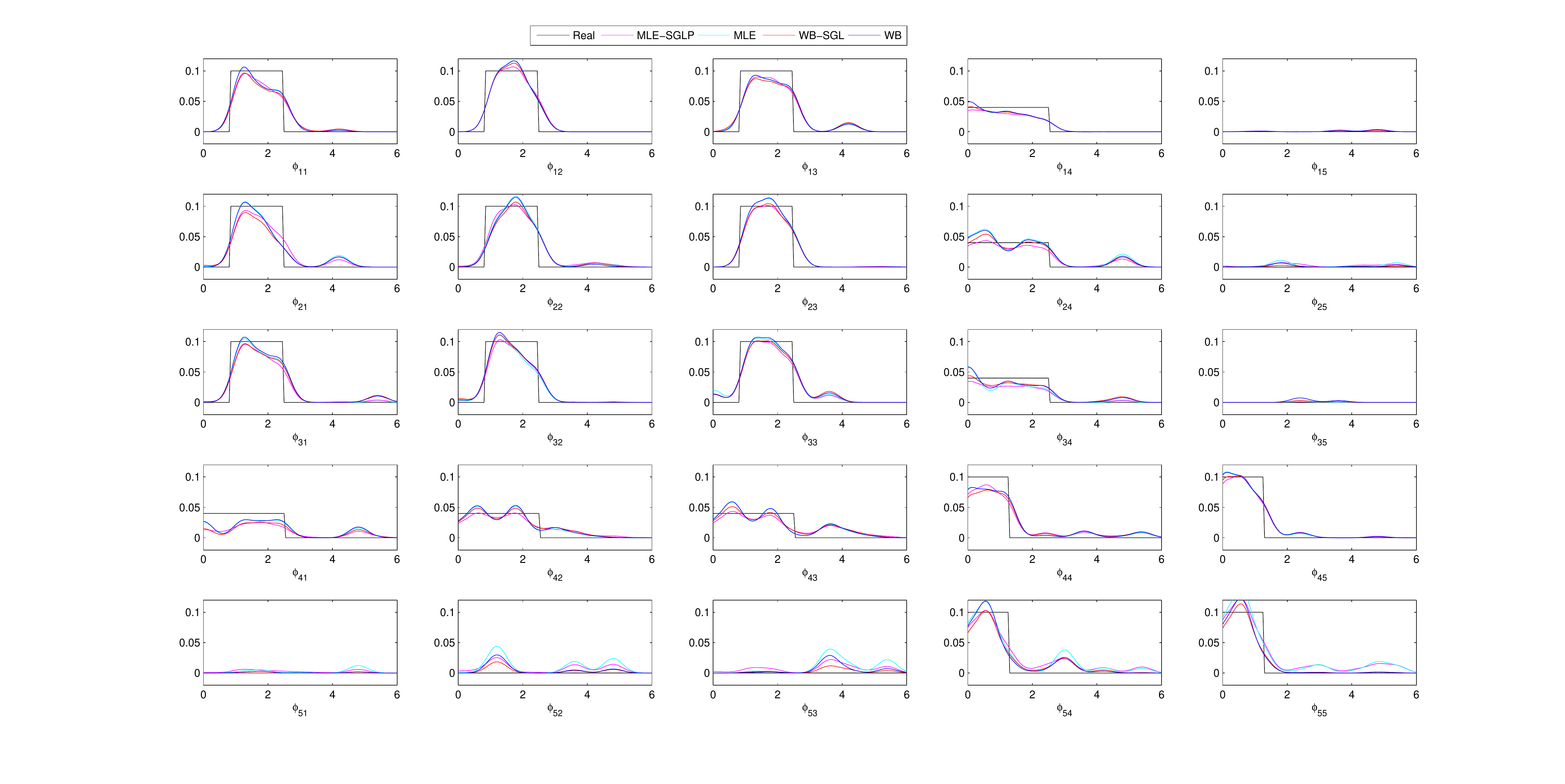}
\end{minipage}%
}%
\centering \caption{The curves of ${\phi _{cc'}}$ of the four sorts
of models: WB,WB-SGL, MLE, and MLE-SGLP }
\end{figure}

Figure 9 illustrates the estimates of the impact functions obtained by these four algorithms. The true values of the function ${\phi _{1,5}}$,${\phi _{2,5}}$ ,${\phi _{3,5}}$ ,${\phi _{5,1}}$ , ${\phi _{5,2}}$and ${\phi _{5,3}}$ are all zero. From Figure 9, we can see that the estimate of the WB-SGL model is the most accurate, which is consistent with the value of ${e_\Phi }(t)$. MLE-based model is misled by the time varying base intensity, due to ${\rho _5} \approx {\rm{1}}{\rm{.24}}$, thus time varying base intensity ${h_5}(t)$ is increasing. Meanwhile, MLE-based model assume that ${h_5}(t)$ is a constant, thus the change of ${h_5}(t)$ is ignored and deteriorate the estimate accuracy of ${\phi _{5c'}}$, reflected in the experimental results, the ${\phi _{5c'}}$ estimate with MLE-based algorithm has a large margin of error. Even the Gaussian basis functions do not fit the square-like impact functions well, our model still can estimate them robustly.
However, the algorithm without regularizer has a worse performance, when the impact functions are all zero, they cannot estimate the impact functions accurately. Regularizers restrict the value of ${a_{cc'm}}$,make the estimate more accurate, but also bring a disadvantage, the restricting the value of ${a_{cc'm}}$  will obtain a lower parameter estimate of ${a_{cc'm}}$ and ${\phi _{cc'}}(t) = \sum\nolimits_1^M {{a_{cc'm}}g(t)} $, the non-zero impact function estimate with regularizers will lower than the ones without regularizers, this point of view can be verified by the experimental results in the Figure 9.

\subsection{Experimental results of constant base intensity data}

In this subsection, to verify the robustness of our algorithms, we test our model on the constant base intensity data \cite{xu2016learning}. In this case, assuming $h(t) = \mu $ will reduce the computational complexity and reduce the interference caused by variable parameters, thus the MLE-based algorithm will have a big advantage. We will still compare all these models on all kinds of measure criteria,such as, ${e_{\bf{\mu }}}$,${e_{\bf{\rho }}}$,${e_{h(t)}}$,${e_\Phi}$ and Loglike.

\subsubsection{Relative error of parameters}

\begin{figure}[htbp]
\centering
\subfigure[Sine-impact function]{
\begin{minipage}[t]{0.5\linewidth}
\centering
\includegraphics[width=0.9\linewidth]{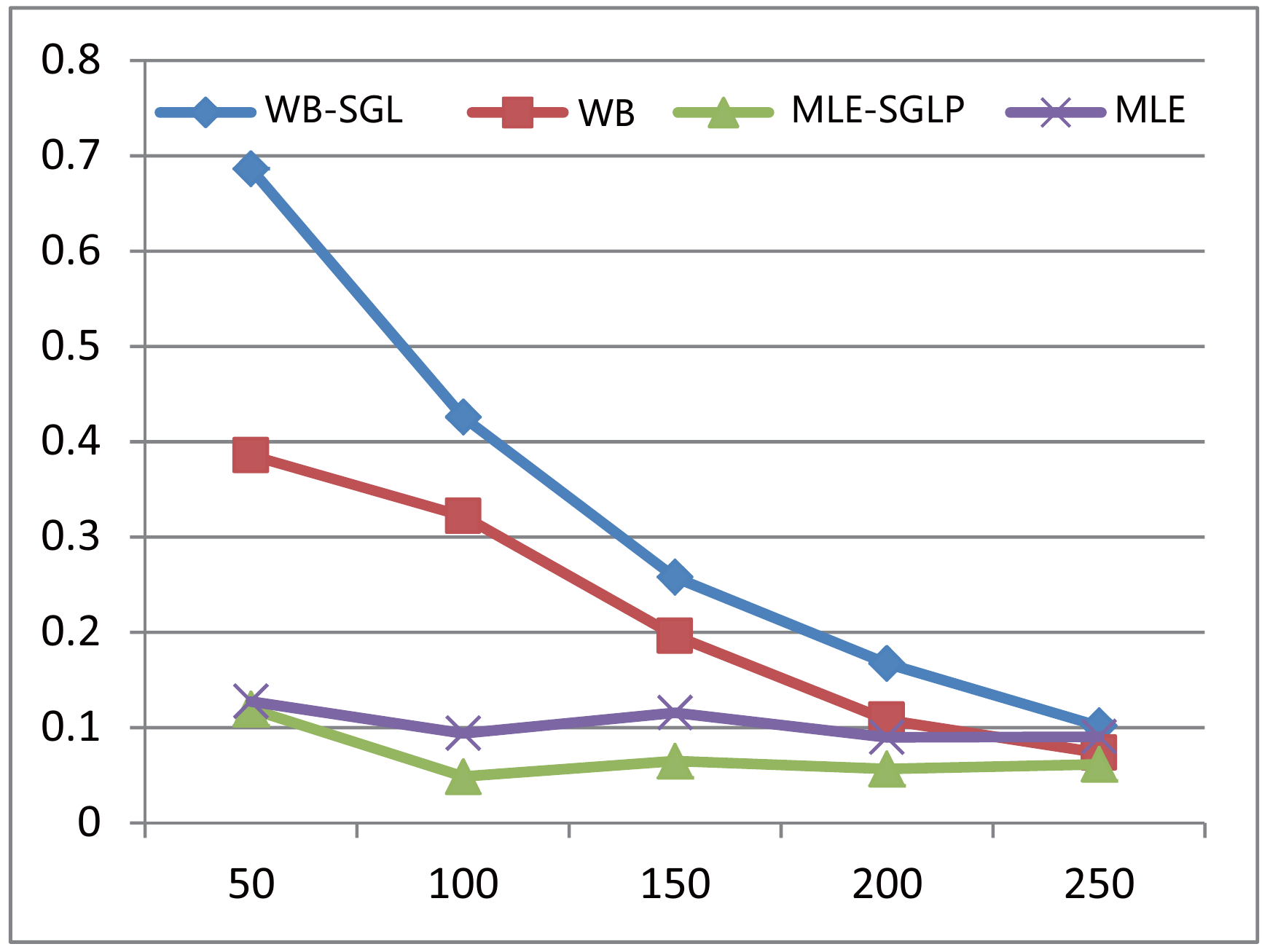}
\end{minipage}%
}%
\subfigure[Square-impact function]{
\begin{minipage}[t]{0.5\linewidth}
\centering
\includegraphics[width=0.9\linewidth]{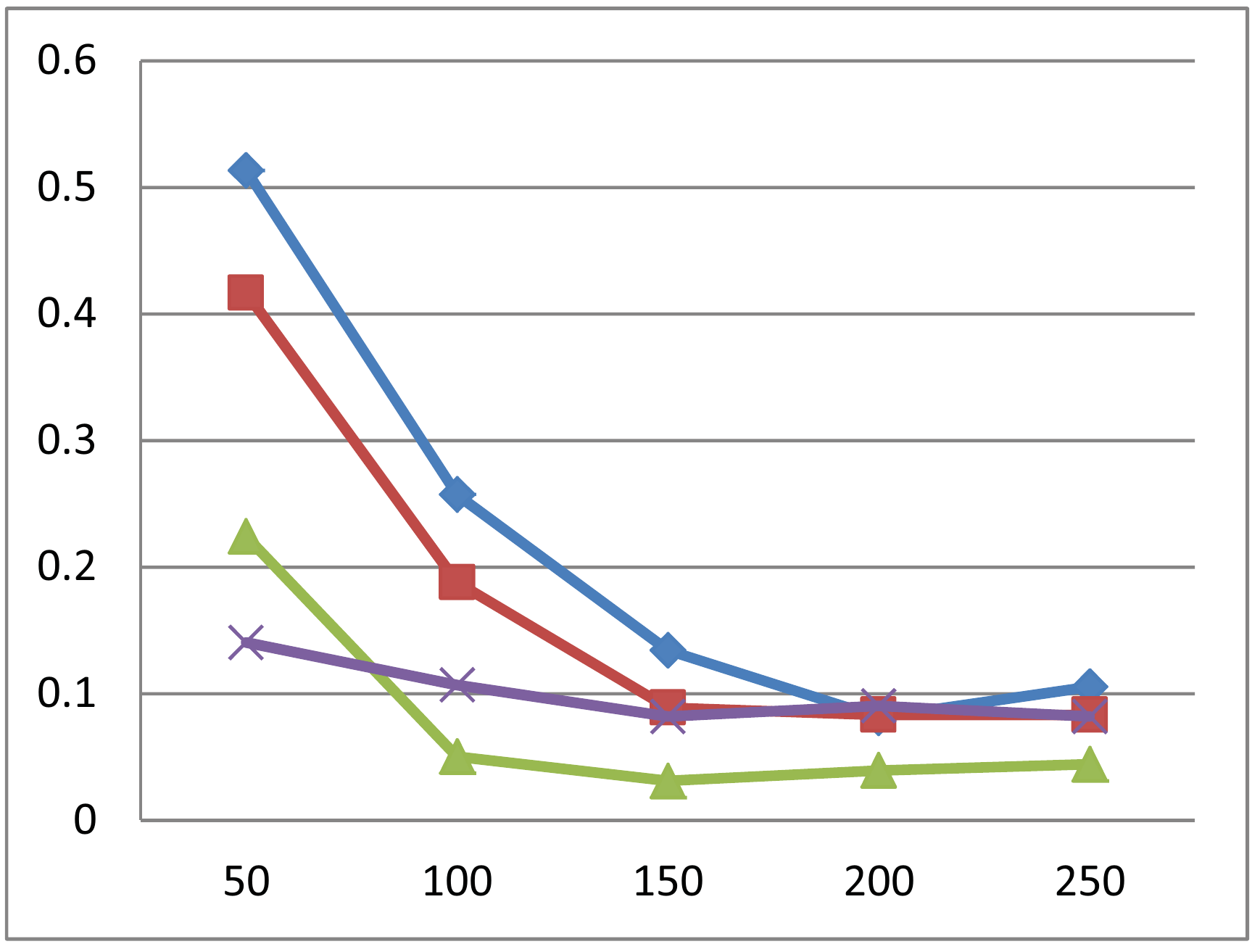}
\end{minipage}%
}%
\centering \caption{The curves of ${e_{\bf{\mu }}}$ of the four sorts of models: WB,WB-SGL, MLE, and MLE-SGLP.}
\end{figure}

In figure 10, we can find that although in WB-based algorithm, we already set the innitial value of  $\rho_c$ to 1, the interference caused by the randomness of the event sequences still affects the estimate of the parameter ${\bf{\mu }}$. However, we can note that as the number of event sequences increases, our algorithm's estimates of ${\bf{\mu }}$ are more and more accuratly. When the number of event sequences is 250, estimate accuracy of our proposed algorithm is basically close to the MLE-based algorithm.

\begin{figure}[htbp]
\centering
\subfigure[Sine-impact function]{
\begin{minipage}[t]{0.5\linewidth}
\centering
\includegraphics[width=0.9\linewidth]{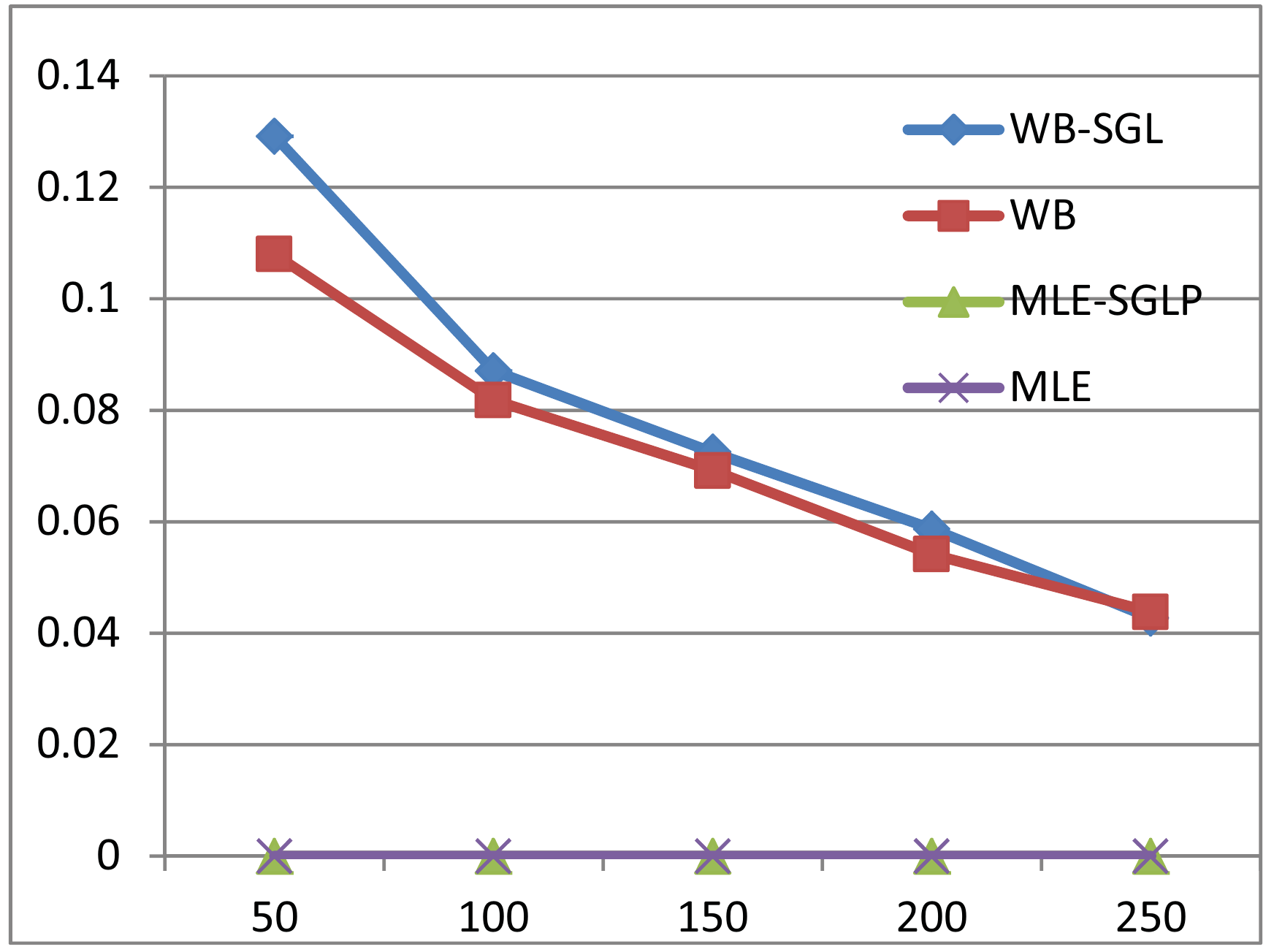}
\end{minipage}%
}%
\subfigure[Square-impact function]{
\begin{minipage}[t]{0.5\linewidth}
\centering
\includegraphics[width=0.9\linewidth]{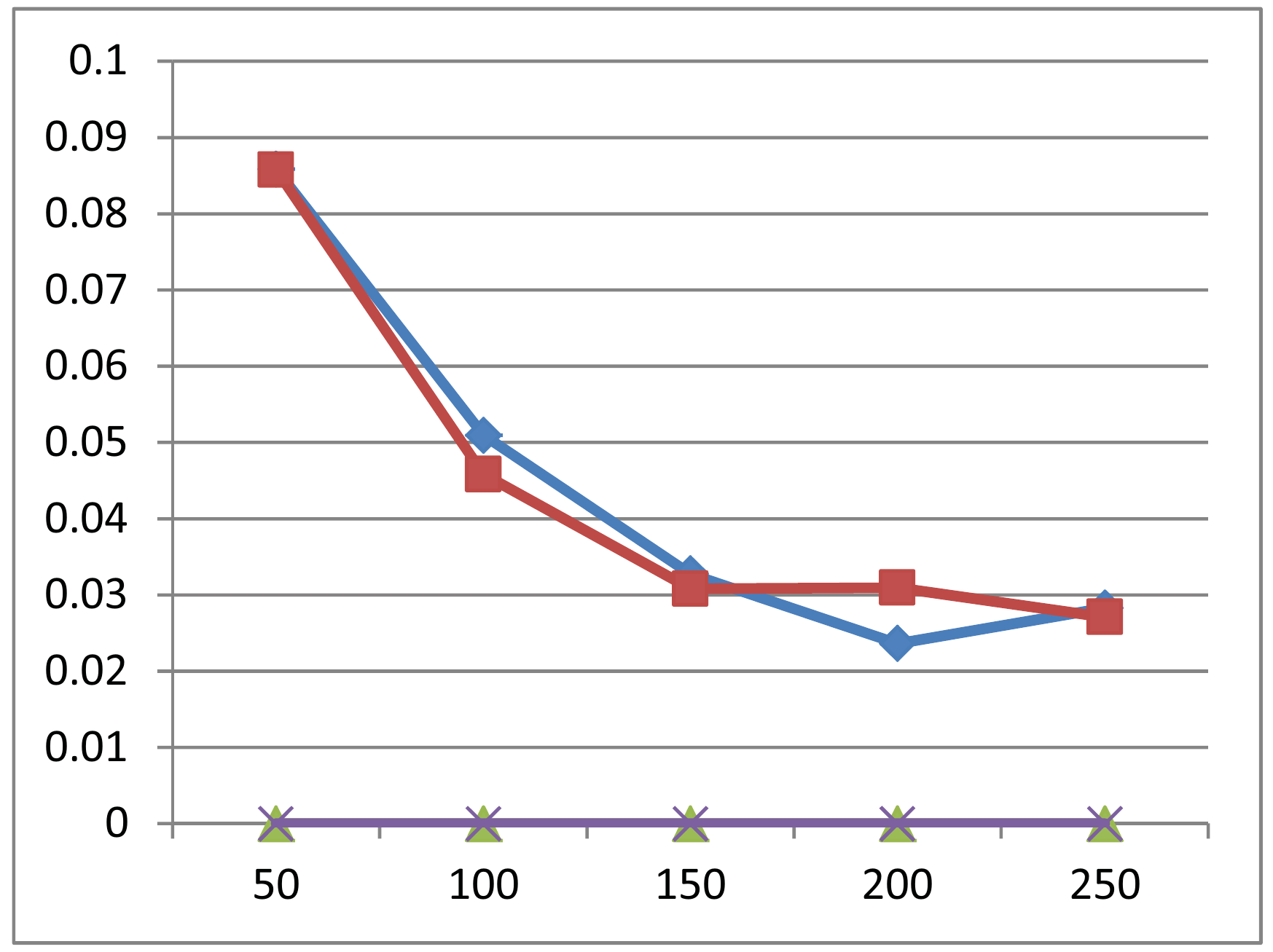}
\end{minipage}%
}%
\centering \caption{The curves of ${e_{\bf{\rho }}}$ of the four sorts of models: WB,WB-SGL, MLE, and MLE-SGLP.}
\end{figure}

In figure 11, we see that MLE-based algorithms assume $\rho_c$ is fixed to 1, thus there is no the estimate error for $\rho_c$ with MLE-based algorithms. Although WB-based algorithms set the innitial value of $\rho_c$  to 1, but due to the randomness of the event sequences, the estimate of $\rho_c$ will fluctuate around the true value. From the experimental results we can get that with increasing the number of event sequences , the estimate of $\rho_c$ is more and more accurate, the relative error is already less than 5\%.

\begin{figure}[htbp]
\centering
\subfigure[Sine-impact function]{
\begin{minipage}[t]{0.5\linewidth}
\centering
\includegraphics[width=0.9\linewidth]{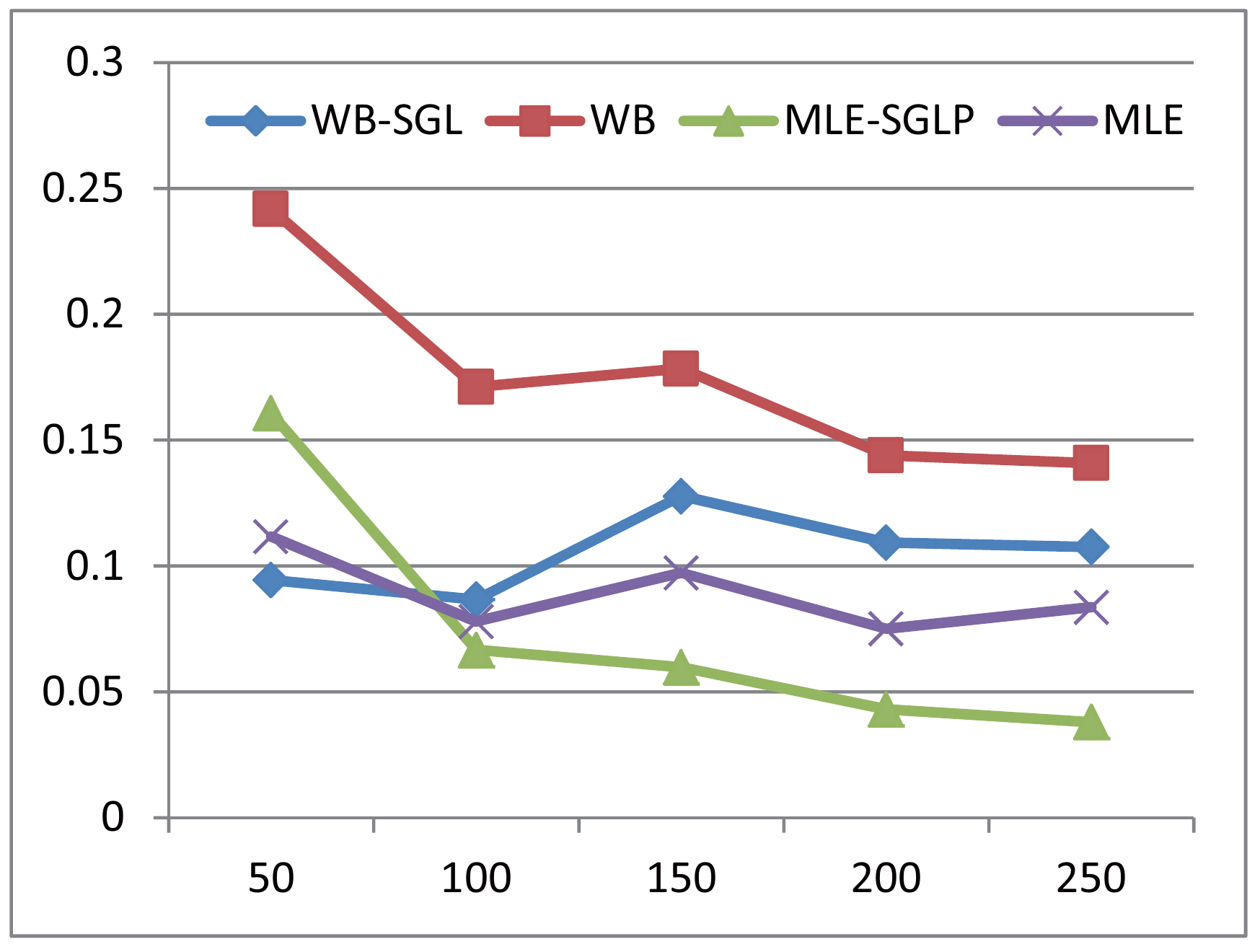}
\end{minipage}%
}%
\subfigure[Square-impact function]{
\begin{minipage}[t]{0.5\linewidth}
\centering
\includegraphics[width=0.9\linewidth]{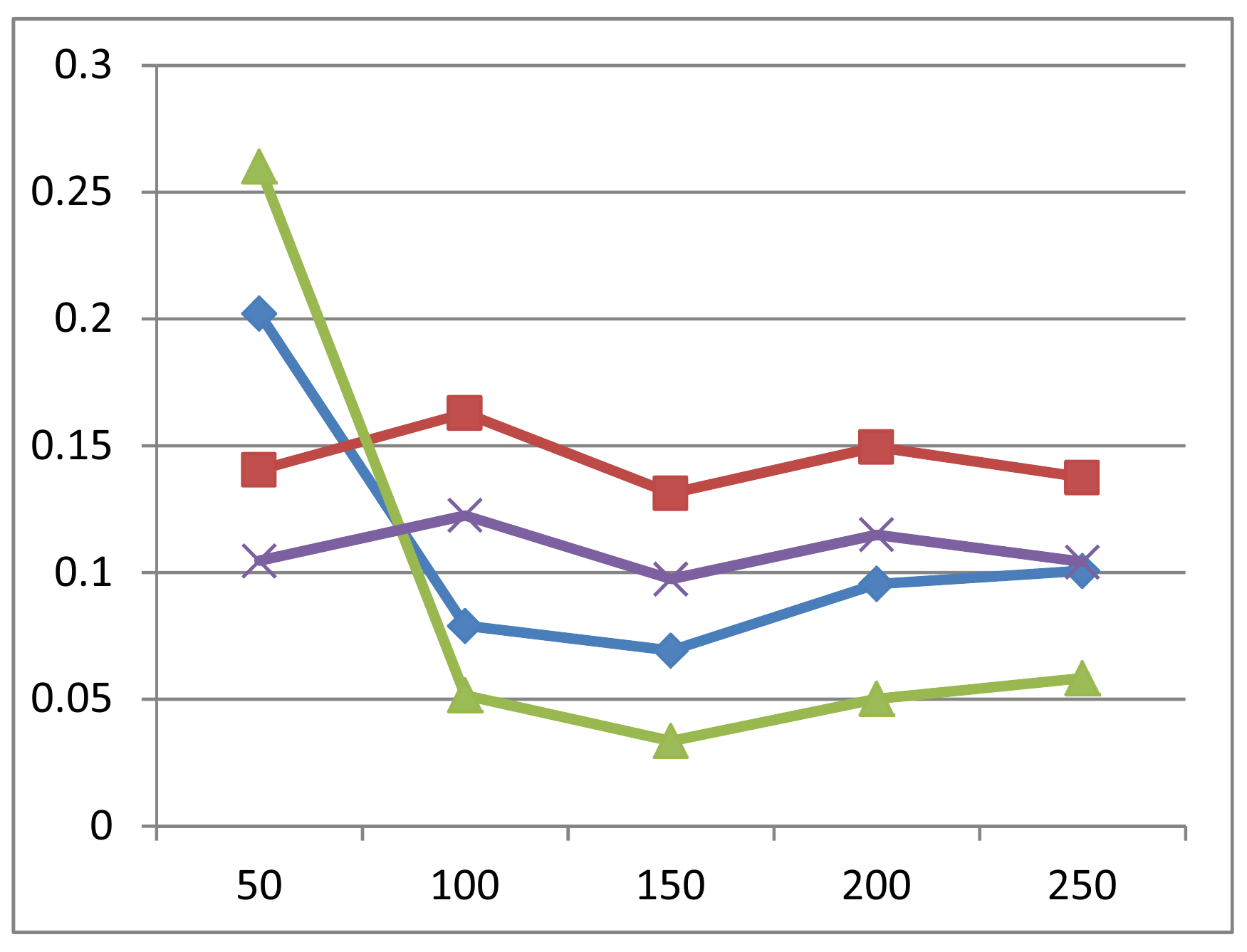}
\end{minipage}%
}%
\centering \caption{The curves of ${e_{h(t)}}$ of the four sorts of models: WB,WB-SGL, MLE, and MLE-SGLP.}
\end{figure}

As showed in figure 12, because of the estimate error of $\bf{\rho }$ and $\bf{\mu}$, the relative error of $h(t)$ with WB-based algorithm is larger than MLE-based algorithm, the relative error with WB-SGL algorithm is closed to the MLE algorithm, which will affect the estimate accuracy of the impact functions.

\begin{figure}[htbp]
\centering
\subfigure[Sine-impact function]{
\begin{minipage}[t]{0.5\linewidth}
\centering
\includegraphics[width=0.9\linewidth]{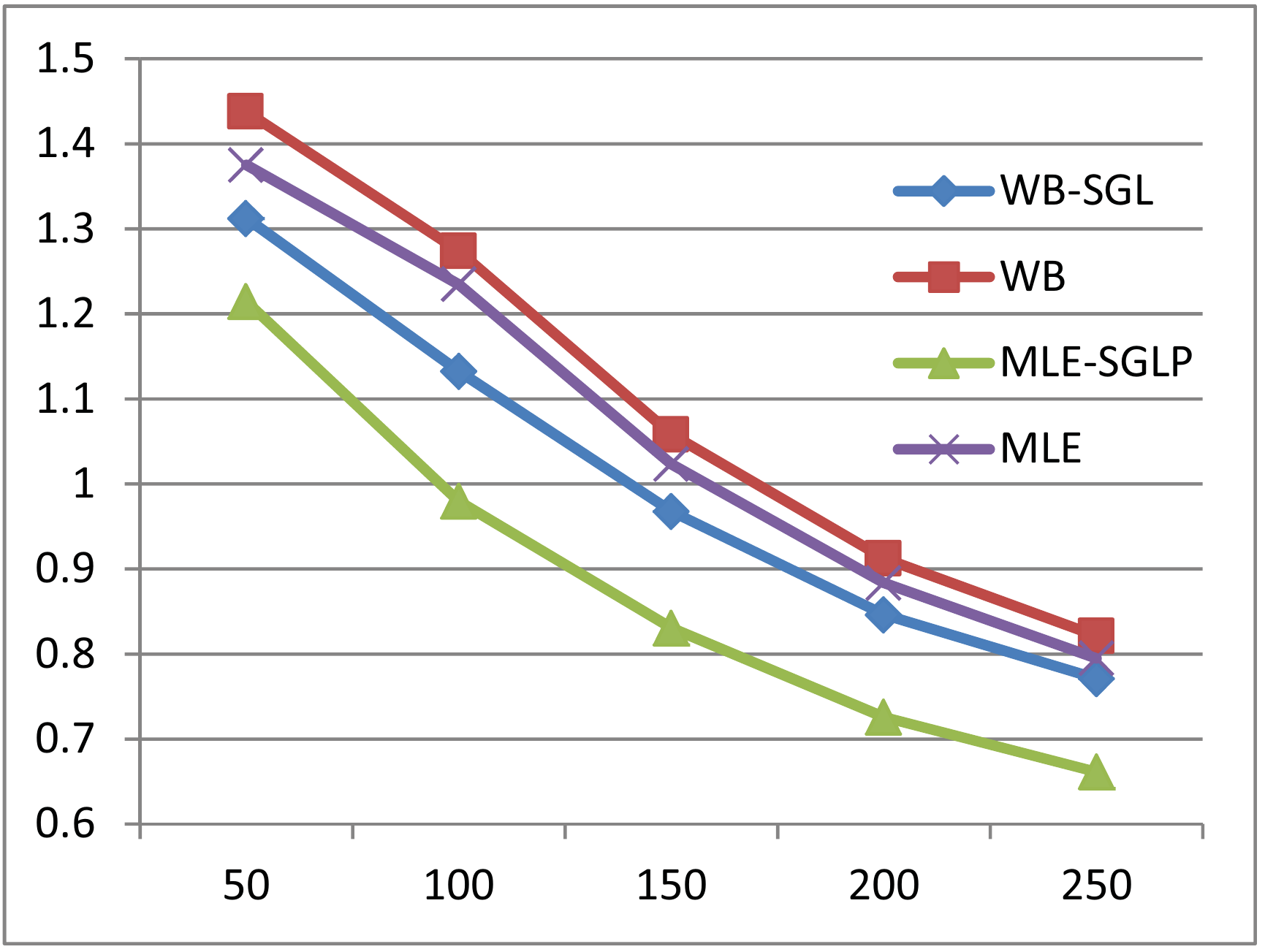}
\end{minipage}%
}%
\subfigure[Square-impact function]{
\begin{minipage}[t]{0.5\linewidth}
\centering
\includegraphics[width=0.9\linewidth]{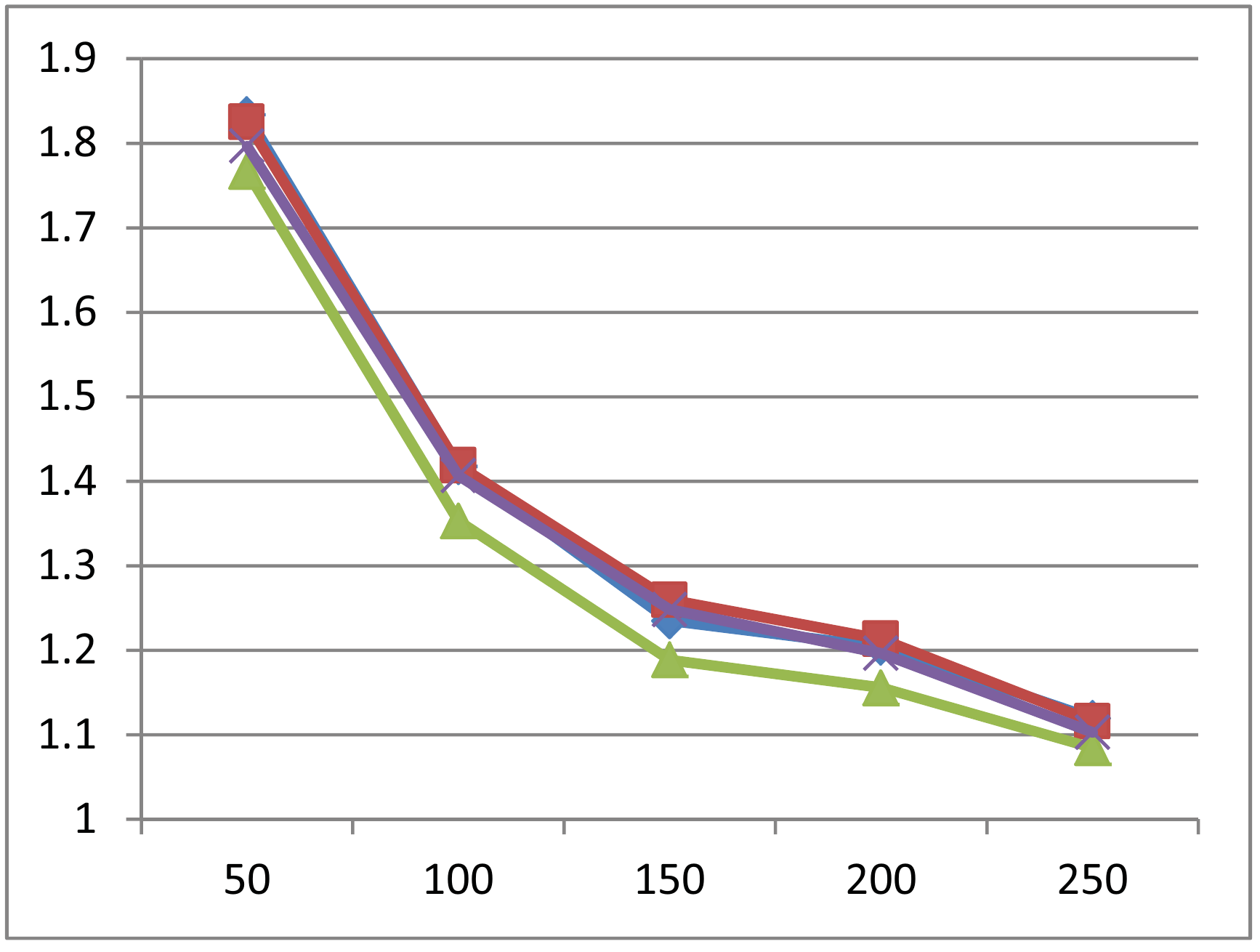}
\end{minipage}%
}%
\centering \caption{The curves of $e_{\Phi}$ of the four sorts of models: WB,WB-SGL, MLE, and MLE-SGLP.}
\end{figure}

\begin{table}
\caption{The ${e_{\Phi (t)}}$ effected by the time varying $h(t)$ (\emph{N}=250)}
\centering
\subtable[WB-based method]{
\begin{tabular}{ccccc}
\hline
                  & \multicolumn{2}{c}{Sine-impact function} & \multicolumn{2}{c}{Square-impact function} \\ \hline
                  & WB-SGL           & WB              & WB-SGL            & WB               \\ \hline
Time varying $h(t)$      & 0.7633           & 0.7811          & 1.1344            & 1.1337           \\
Constant    $h(t)$        & 0.7712           & 0.8216          & 1.1197            & 1.1157           \\ \hline
\end{tabular}
       \label{tab:firsttable}
}

\quad
\quad

\subtable[MLE-based method]{
\begin{tabular}{ccccc}
\hline
             & \multicolumn{2}{c}{Sine-impact function} & \multicolumn{2}{c}{Square-impact function} \\ \hline
             & MLE-SGLP          & MLE            & MLE-SGLP           & MLE             \\ \hline
Time varying $h{(t)}$ & 0.7729            & 0.8983         & 1.2184             & 1.265           \\
Constant   $h{(t)}$  & 0.6613            & 0.795          & 1.0846             & 1.103           \\ \hline
\end{tabular}
       \label{tab:secondtable}
}
\end{table}

From the Figure 13 and Table 1, we can find out that our proposed algorithm is more robust than MLE-based algorithm,the varying range of the estimate error of MLE-based alogrithm's impact function ${e_{\Phi (t)}}$, is quite larger than WB-based algorithm, although in constant intensity event sequence, MLE-SGLP model has the most accurate estimate of $\Phi (t)$. There is another very interesting findings.In square-like case , the porformance of WB-based algorithm in constant $h(t)$ event sequence is a little bit better than in time varying $h(t)$ event sequence. In our view, the reason for this phenomenon is that, in the case of constant base intensity, the assumption that the base intensity is time-varying compensates for incompatibility between gaussian basis functions ${g_m}(t)$ and square-like impact functions ${\phi _{cc'}}(t)$.

\begin{figure}[htbp]
\centering
\subfigure[Sine-impact function]{
\begin{minipage}[t]{0.5\linewidth}
\centering
\includegraphics[width=0.9\linewidth]{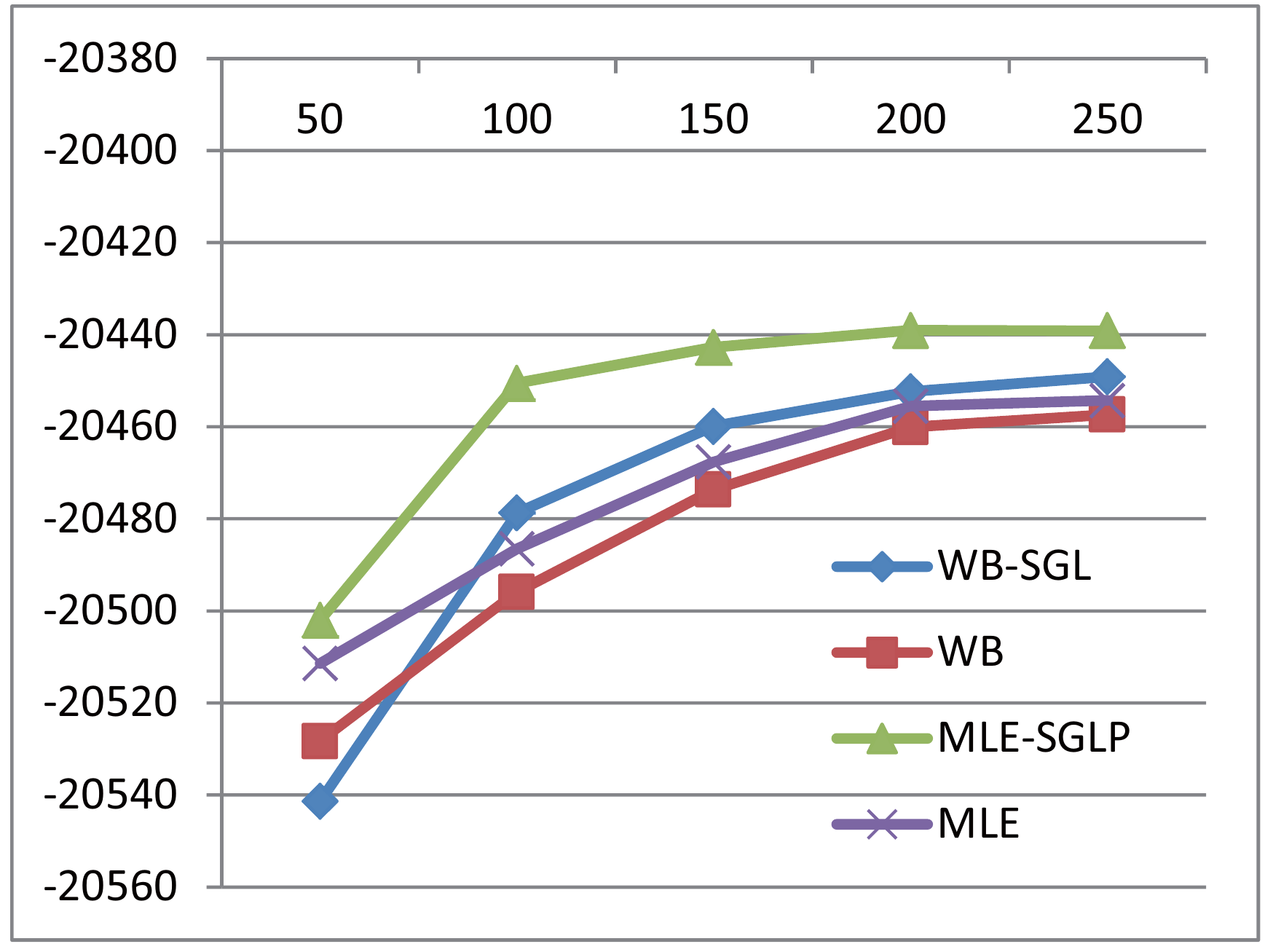}
\end{minipage}%
}%
\subfigure[Square-impact function]{
\begin{minipage}[t]{0.5\linewidth}
\centering
\includegraphics[width=0.9\linewidth]{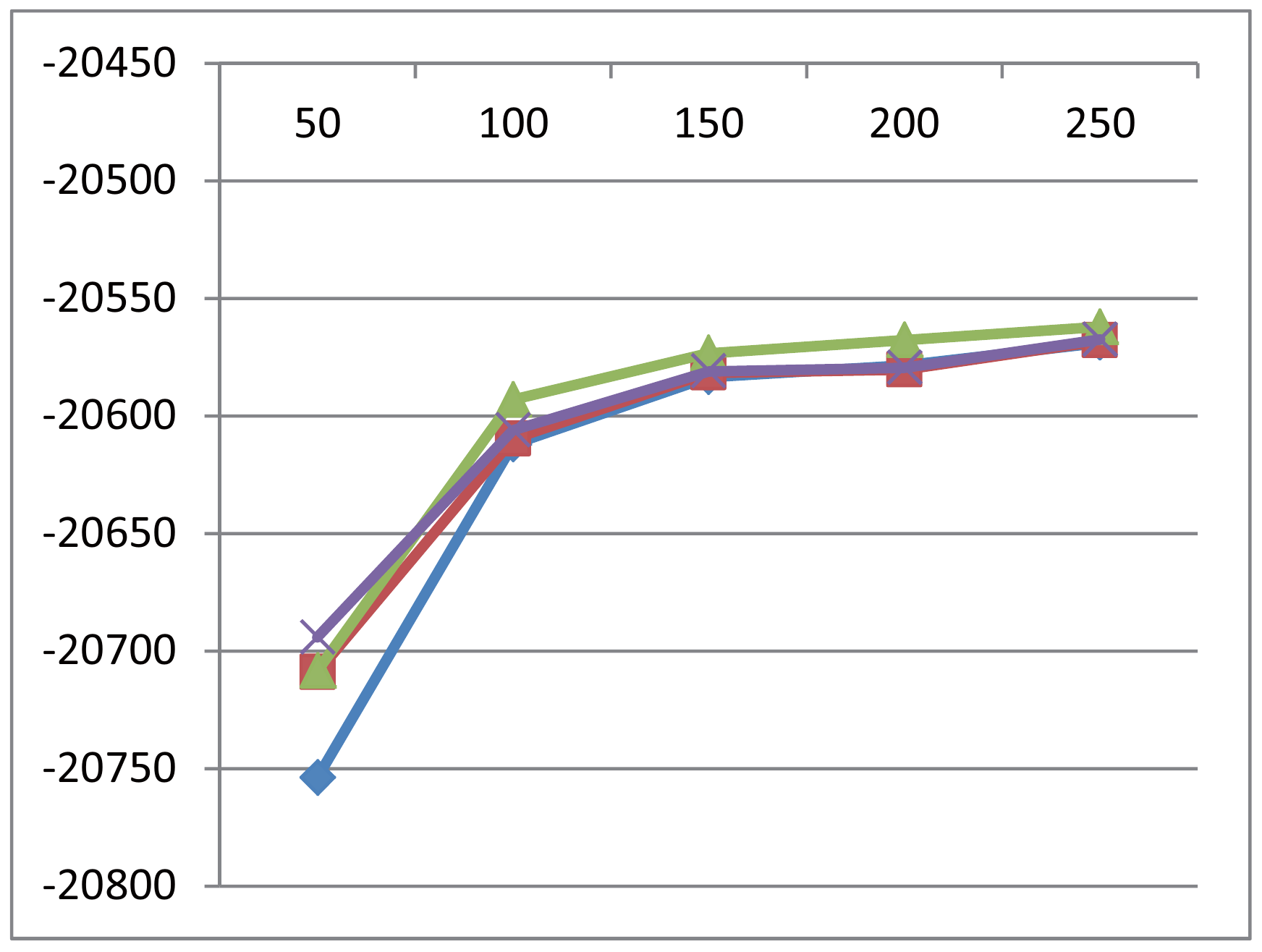}
\end{minipage}%
}%
\centering \caption{The curves of Loglike of the four sorts of models: WB,WB-SGL, MLE, and MLE-SGLP.}
\end{figure}

Based on the above discussions, we will analysis Loglike of different models. Undoubtly, from Figure 14, we can find that the Loglike of MLE-SGLP model is the highest than other models, which verifies that MLE-SGLP model is superior to other models, and WB-SGL model is superior to pure MLE model. However, the most important point is that the difference between the Loglike of those algorithms is smaller than in time-varying $h(t)$  event sequences. At the same time, the analysis results of the relative errors of other parameters can also verify this, such as ${e_{\bf{\mu }}}$, ${e_{\bf{\rho }}}$,${e_{h(t)}}$ ,and ${e_{\bf{\Phi}}}$. This validates that the WB-based algorithm we proposed have strong stability and robustness, both on the time-varying $h(t)$  event sequences and constant $h(t)$  event sequences, WB-based algorithm can obtain more accurate parameter estimation and have stronger generalization ability.

\subsubsection{Curves of the impact functions}

\begin{figure}[htbp]
\centering \subfigure[sine-impact function]{
\begin{minipage}[t]{1.0\linewidth}
\centering
\includegraphics[trim = 40mm 0mm 0mm 0mm,width=1.0\textwidth]{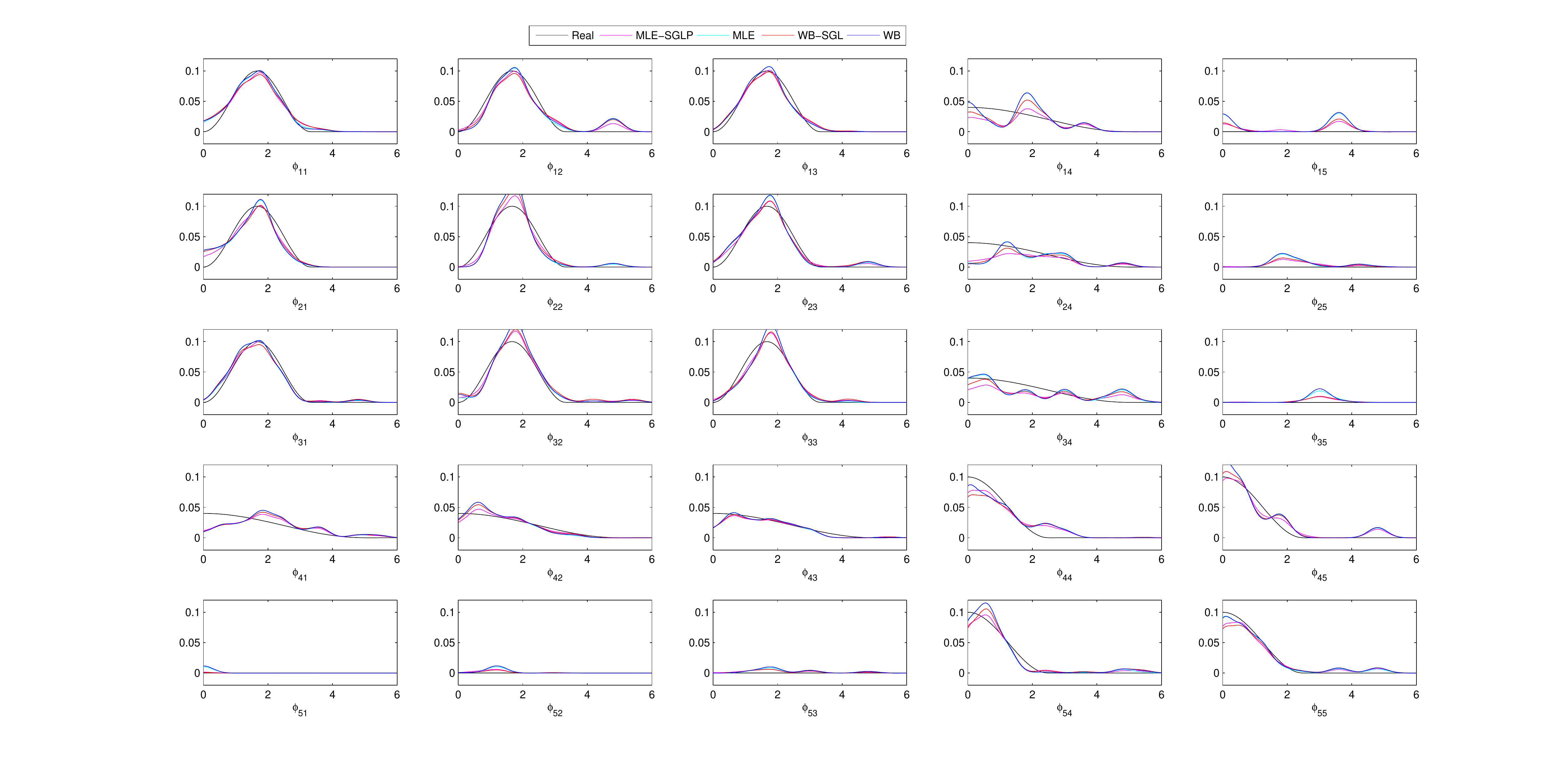}
\end{minipage}%
}%
\quad \subfigure[square-impact function]{
\begin{minipage}[t]{1.0\linewidth}
\centering
\includegraphics[trim = 40mm 0mm 0mm 0mm,width=1.0\textwidth]{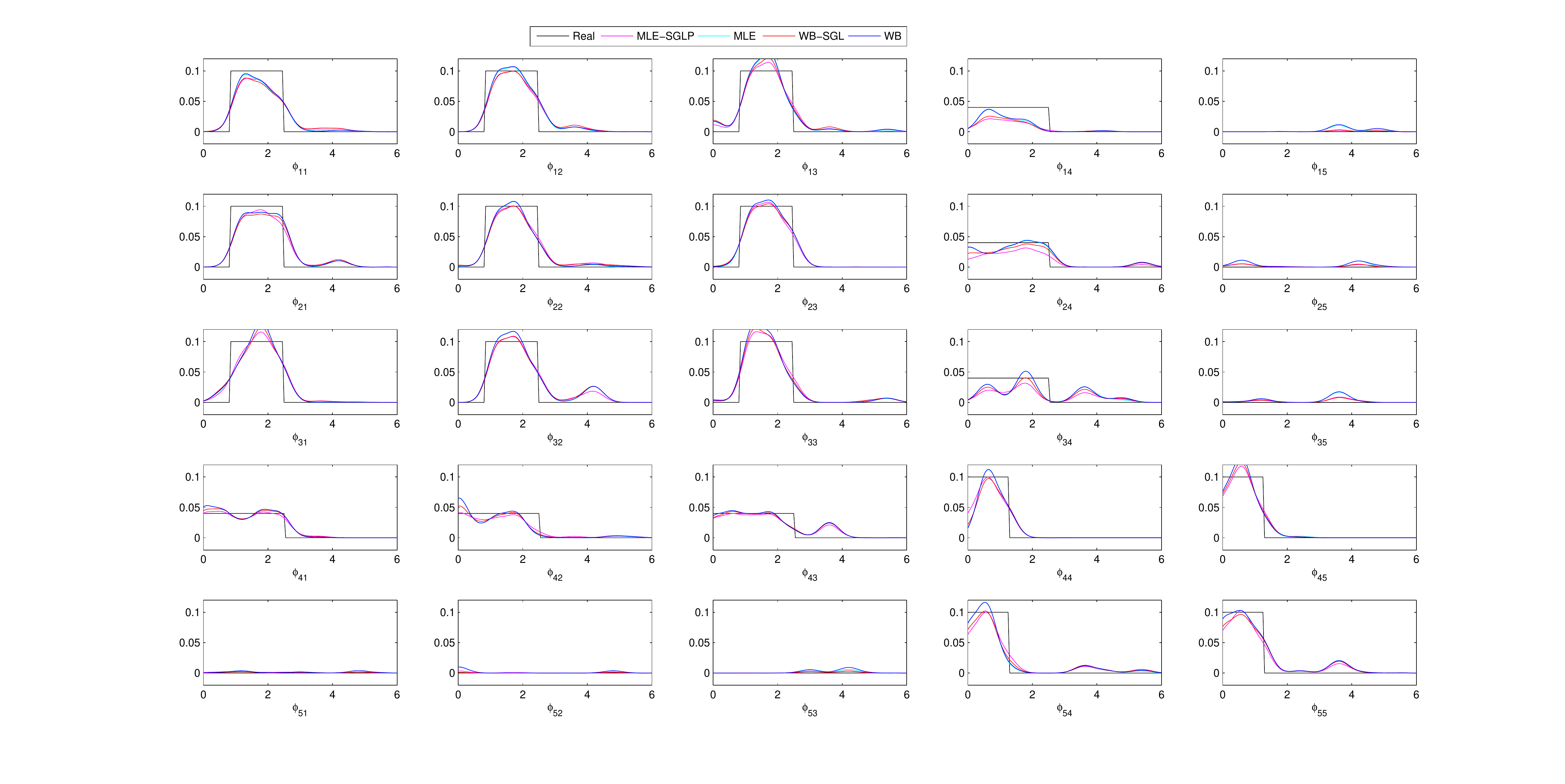}
\end{minipage}%
}%
\centering \caption{The curves of ${\phi _{cc'}}$ of the four sorts
of models: WB,WB-SGL, MLE, and MLE-SGLP }
\end{figure}

Figure 15 illustrates the estimates of the impact functions obtained by these four algorithms on constant $h(t)$ event sequences. The algorithms with regularizers, such as MLE-SGLP and WB-SGL algorithm are more accuracy in finding the Granger causality. Pure WB algorithm's performance is worst, because of the estimation error of $h(t)$ and without using the regularizers for $a_{cc'm}$ .

\subsection{Real-world data}

In this section, we will verify the validity of our proposed model on real-world data. This data is labeled by us and is named compressor station failure data set. And on this data set we set ${\alpha _S} = 1$ and ${\alpha _G} = 1$. The data set contains 87 compressor stations'  failure event sequences from March, 2006 to October, 2018. There are 14 subsystems in each compressor station, so all the failures are categorized into 14 categories as Table 2 according to the subsystem in which the failure occurred.

\begin{table}[htbp]
    \centering
    \caption{The 14 types of failures}
    \begin{tabular}{cc}
        \toprule  
        Index&Failure types \\
        \midrule  
        1&Gas Generator system failure \\
        2&Motor system failure \\
        3&Engine system failure \\
        4&Power supply system failure \\
        5&Fuel gas system failure \\
        6&Lubricating oil system failure \\
        7&Intake system failure\\
        8&Control system failure \\
        9&Cooling system failure \\
        10&Firefighting system failure \\
        11&Compressor system failure \\
        12&Instrument air system failure\\
        13&Process system failure \\
        14&Other failure\\
        \bottomrule  
    \end{tabular}
\end{table}

All of these types of failures have both self-triggering and mutually- triggering patterns. For instance, if the engine system fails, the possibility of future engine system failure will increase (self-triggering), if the gas generator system fails, the possibility of future control system failure will increase (mutually-triggering). Not only that, the trend of the base intensity of each system failure with time can also be obtained through learning which is depending on the parameters $\bm{\rho}$ and $\bm{\mu}$ .Thus, we model the failure event sequence via a Hawkes process with time-varying base intensity, which is from Weibull distribution, then we can learn the Granger causality among the failure categories and the varying trend of base intensity.

\subsubsection{The Loglike of different models}

\begin{table}[]
\caption{The Loglike of various models on real-world data}
\resizebox{\textwidth}{15mm}{
\begin{tabular}{cccccccc}
\hline
\multicolumn{4}{c}{WB series model}    & \multicolumn{4}{c}{MLE series model}  \\ \hline
WB-SGL  & WB-S    & WB-GL    & WB      & MLE-SGL & MLE-S   & MLE-GL  & MLE     \\
-5054.8 & -5055.8 & -5056.21 & -5061.3 & -5755.3 & -5610.3 & -5760.7 & -5764.5 \\ \hline
\end{tabular}}
\end{table}

The sequence of fault events of eighty compressor stations is used as training data, and the sequence of fault events of the remaining compressor stations is used as test data. Consider the fact that the compressor station is inspected for each production quarter and the shortest stable running time is three days, we set the length of time of the impact function to be 90 days (the influence of a failure will not exist in a production quarter), and the number of basis function $M=31$ (sampling every 72 hours). The parameter of the kernel Gaussian function is defined as:

$$\sigma {\rm{ = }}{{{\rm{90}}} \over {30 \times 2}}{\rm{ = }}1.5$$

Table 3 lists the Loglike obtained via 8 sorts of models (WB-SGL, WB-S, WB-GL, WB, MLE-SGL, MLE-S, MLE-GL, and MLE) with compressor station failure data. We can figure out that the WB model is far better than the MLE model, and WB-SGL model obtains the best experimental result. This is in line with the assumptions we made at the beginning of this paper, the base intensity of the Hawkes process is not a constant in practical applications, but a time-varying function. And we assume that the base intensity is consistent with the Weibull distribution and it has achieved better results in practical applications.

\subsubsection{The estimation of Weibull base intensity parameters}
\begin{table}[]
\caption{The parameter estimation of various models on real-world data}
\resizebox{\textwidth}{50mm}{
\begin{tabular}{ccccccccc}
\hline
Index     & \multicolumn{2}{c}{WB-SGL} & \multicolumn{2}{c}{WB-S} & \multicolumn{2}{c}{WB-GL} & \multicolumn{2}{c}{WB} \\ \hline
Parameter &   $ \mu $     &  $\rho$    &  $ \mu $     &   $\rho$  &    $ \mu $   &   $\rho$   &   $ \mu $   &   $\rho$ \\ \hline
1         & 0.004168      & 0.676046   & 0.004158     & 0.676254  & 0.004166     & 0.676083   & 0.004164    & 0.676103 \\
2         & 0.002214      & 0.69372    & 0.002216     & 0.693572  & 0.002219     & 0.69345    & 0.002219    & 0.69341  \\
3         & 0.000413      & 0.894198   & 0.000412     & 0.894285  & 0.000413     & 0.894166   & 0.000412    & 0.894279 \\
4         & 0.003276      & 0.886825   & 0.003273     & 0.88689   & 0.003276     & 0.886823   & 0.003273    & 0.886886 \\
5         & 0.000433      & 0.907548   & 0.000432     & 0.907589  & 0.000433     & 0.907548   & 0.000432    & 0.907592 \\
6         & 0.001012      & 0.810709   & 0.001011     & 0.810765  & 0.001013     & 0.810642   & 0.001011    & 0.810756 \\
7         & 0.000263      & 0.940528   & 0.000262     & 0.940535  & 0.000263     & 0.940528   & 0.000262    & 0.940557 \\
8         & 0.01711       & 0.763604   & 0.017102     & 0.763622  & 0.017122     & 0.763525   & 0.017098    & 0.763646 \\
9         & 0.001053      & 0.779977   & 0.001052     & 0.780047  & 0.001054     & 0.779939   & 0.001053    & 0.779989 \\
10        & 9.08$ \times {10^{ - 5}}$   & 0.980752   & 9.07$ \times {10^{ - 5}}$   & 0.980747  & 9.08$ \times {10^{ - 5}}$  & 0.980753   & 9.07$ \times {10^{ - 5}}$  & 0.980745 \\
11        & 0.0129        & 0.595825   & 0.012873     & 0.596012  & 0.012892     & 0.595895   & 0.012874    & 0.596006 \\
12        & 0.008312      & 0.433294   & 0.00829      & 0.43353   & 0.008317     & 0.433229   & 0.008294    & 0.433486 \\
13        & 0.000224      & 0.895607   & 0.000223     & 0.895686  & 0.000224     & 0.895594   & 0.000224    & 0.895627 \\
14        & 1.85$ \times {10^{ - 5}}$   & 0.985088   & 1.85$ \times {10^{ - 5}}$   & 0.9851    & 1.85$ \times {10^{ - 5}}$   & 0.985088   & 1.85$ \times {10^{ - 5}}$ & 0.985104 \\ \hline
\end{tabular}}
\end{table}

In order to get the most accuracy estimation of the parameters, we learn our model on the all failure event sequence. Table 4 shows the estimates of $\mu$ and $\rho$ via four different WB-based models. We can see that the parameter estimates of different models are basically the same, which can verify that our algorithm is stable. Below we analyze the reliability of the compressor station with the parameters of the best model on test results, i.e., the WB-SGL model.

Based on the estimates of parameters $\mu$ and $\rho$ with WB-SGL model, we can figure out that these compressors stations are in normal working condition and have superior reliability. However, there are still two issues that need our attention. For instance, Control system, Cooling system, Compressor system and Instrument system failure have a higher value of scale parameter, and a lower value of shape parameter $\rho$ , according to the definition of hazard function of Weibull base intensity $h(t) = \mu \rho {t^{\rho  - 1}}$ , these failures are more likely to occur when the gas station is just starting production, and the likelihood of these occurred failures will gradually decrease over time.

On the other hand, for Fuel gas system failures, Intake system failures, Firefighting system failures and other failures, due to the high value of $\rho$ ($\rho>0.9$), although it is still less than 1, the parameter estimation error, the environmental negative affect, misoperation, and equipment aging, etc., may gradually shift the hazard function from decreasing over time to increasing over time, thus we need to pay attention to the risks of these failures after a long run. The exciting thing is that the estimated trend of system reliability based on parameters is consistent with the experiences of the operation and maintenance experts of the compressor station. This proves the validity and accuracy of our proposed algorithm.

\subsubsection{Granger causality of different models}

In this subsection, we will analyze the trigger patterns of all kinds of failures. We test the four models both on our model and the MLE model on the entire data set. In order to more intuitively represent the causal relationship between each type of fault, we construct the infectivity matrixes of different types of failure in Figure 16. The element in the $c$-th row and the $c'$-th column of the infectivity matrixes is $\int_0^\infty  {{\phi _{cc'}}(s)ds} $ . The infectivity matrixes of the same series of models are basically the same, so here we only show the infectivity matrixes of the WB-SGL and the MLE-SGL model.

\begin{figure}[htbp]
\centering \subfigure[MLE-SGL]{
\begin{minipage}[t]{0.5\linewidth}
\centering
\includegraphics[width=1.0\textwidth]{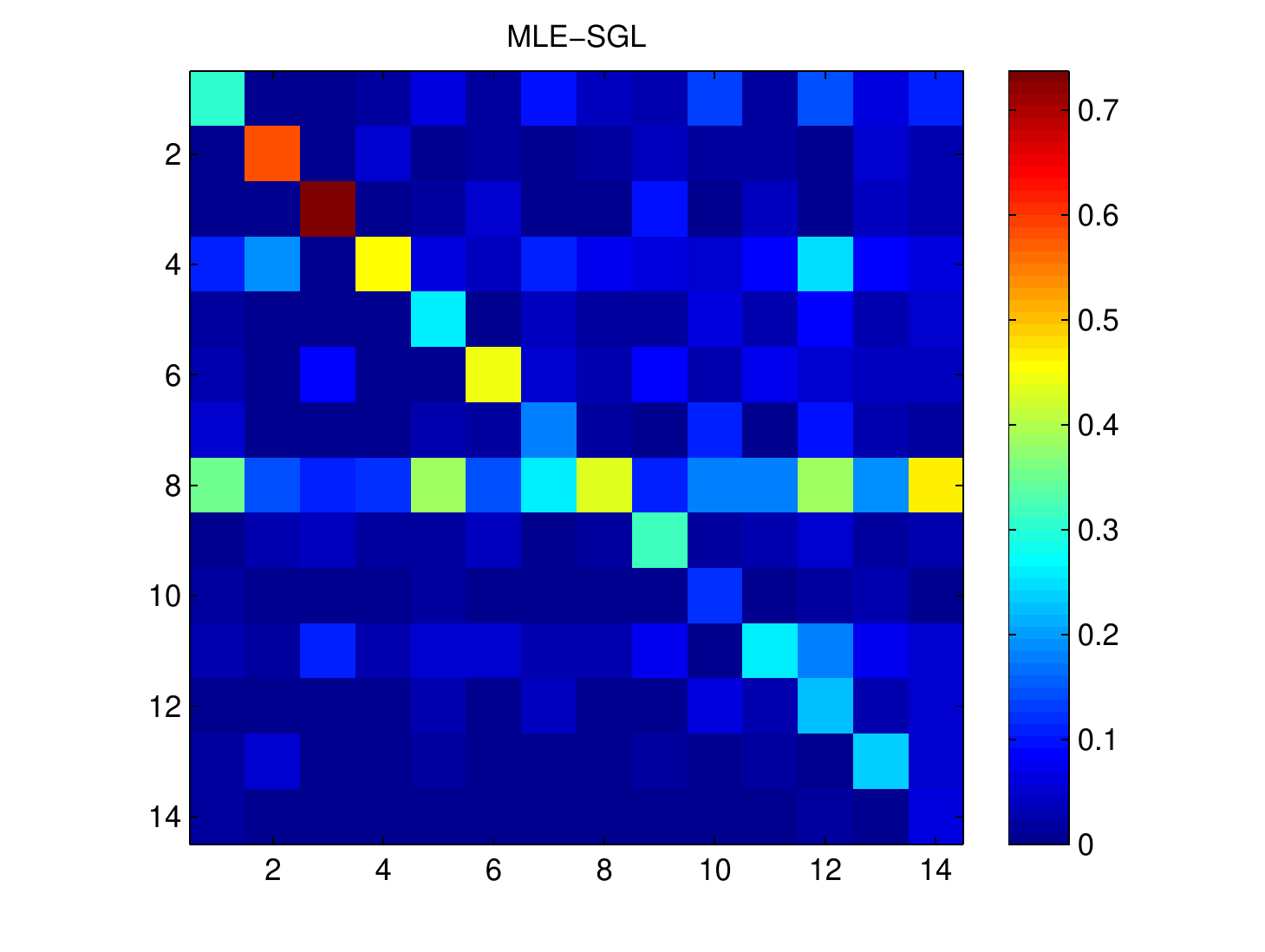}
\end{minipage}%
}%
\subfigure[WB-SGL]{
\begin{minipage}[t]{0.5\linewidth}
\centering
\includegraphics[width=1.0\textwidth]{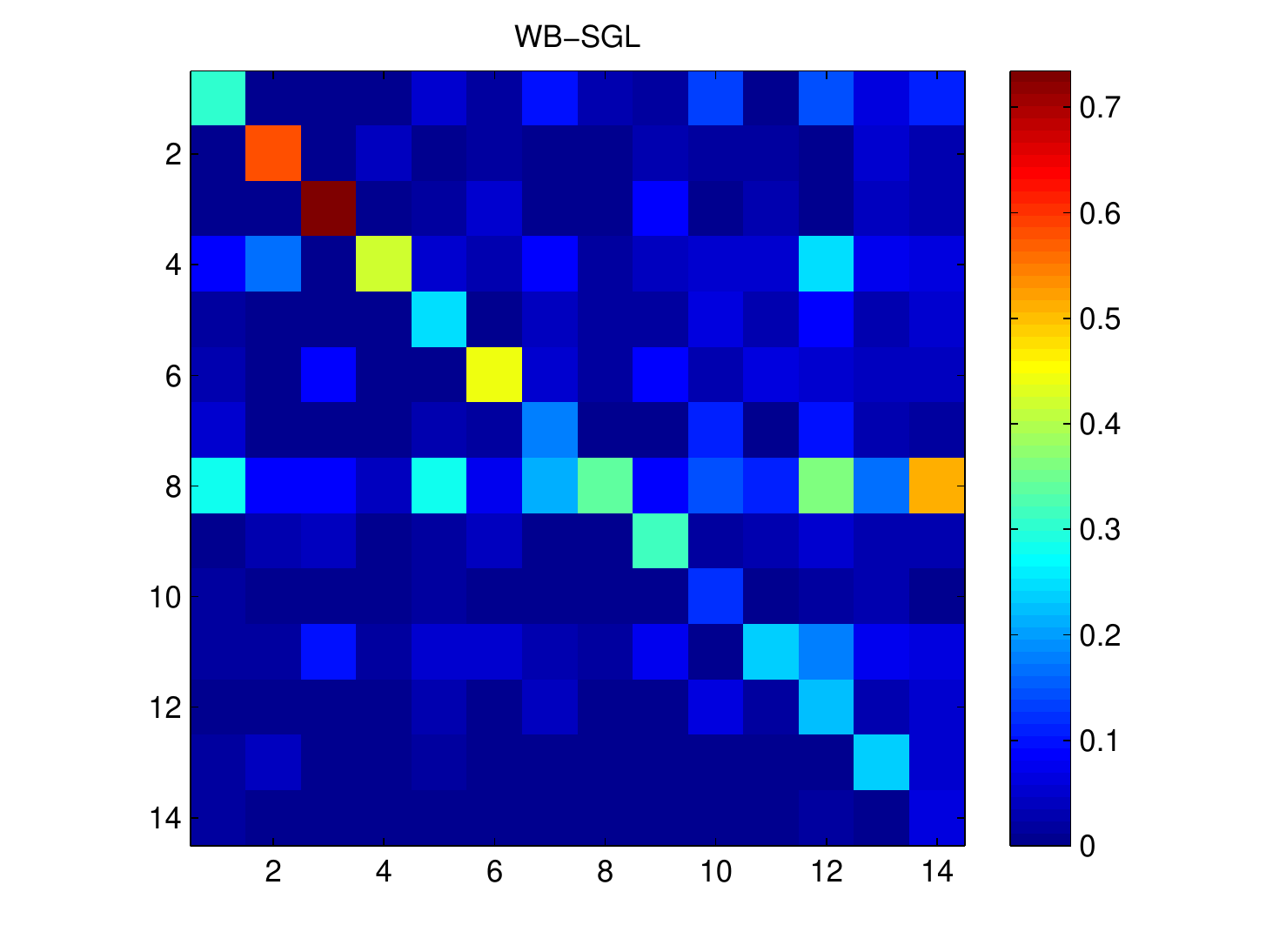}
\end{minipage}%
}%
\centering \caption{The infectivity matrixes of WB-SGL and MLE-SGL}
\end{figure}

The closer the color in the matrix is to blue, which indicates the weaker the causal relationship between the two types of faults. We can see that most of the elements in most infectivity matrices are dark blue, which reflects the sparseness of Granger causality. From Figure 16, It can be seen that the sparsity of our proposed model is better than that of the MLE-SGL model.

In general, after a certain type of fault occurs, the probability of occurrence of this fault will increase. This means that after a certain type of fault occurs, there is a possibility of a secondary failure in a short period of time. The infectivity matrixes reveal this phenomenon: the main diagonal elements of the infectivity matrixes are larger than most other elements in the matrixes. Especially in engine system, motor system, power supply system, and lubricant system's diagonal elements are larger than other main diagonal elements. This result implies that Engine system failure has the strongest self-trigger correlation, which implies that if an engine system failure occurred then the probability of a secondary failure occurring within 90 days is quite large. Motor system, power supply system, and lubricant system are also prone to secondary failures.

In addition, after other systems fails, the control system is most susceptible to failures caused by remain system failures, such as gas generator system failure, power supply system failure, instrument air system failure and other failure. This fact reflects the susceptibility of the control system, because there are intimate relations between control system and other system.

For a more detailed analysis of Granger causality between failures and impact function (which indicates failure probability under the influence of historical fault events) over time, we rank the infectivity element from high to low, and show the top-28 impact functions in Fig 17. Observing top-28 impact functions, we can roughly divide the impact function into three categories.

\begin{figure}
\begin{center}
\includegraphics[width=1.0\textwidth]{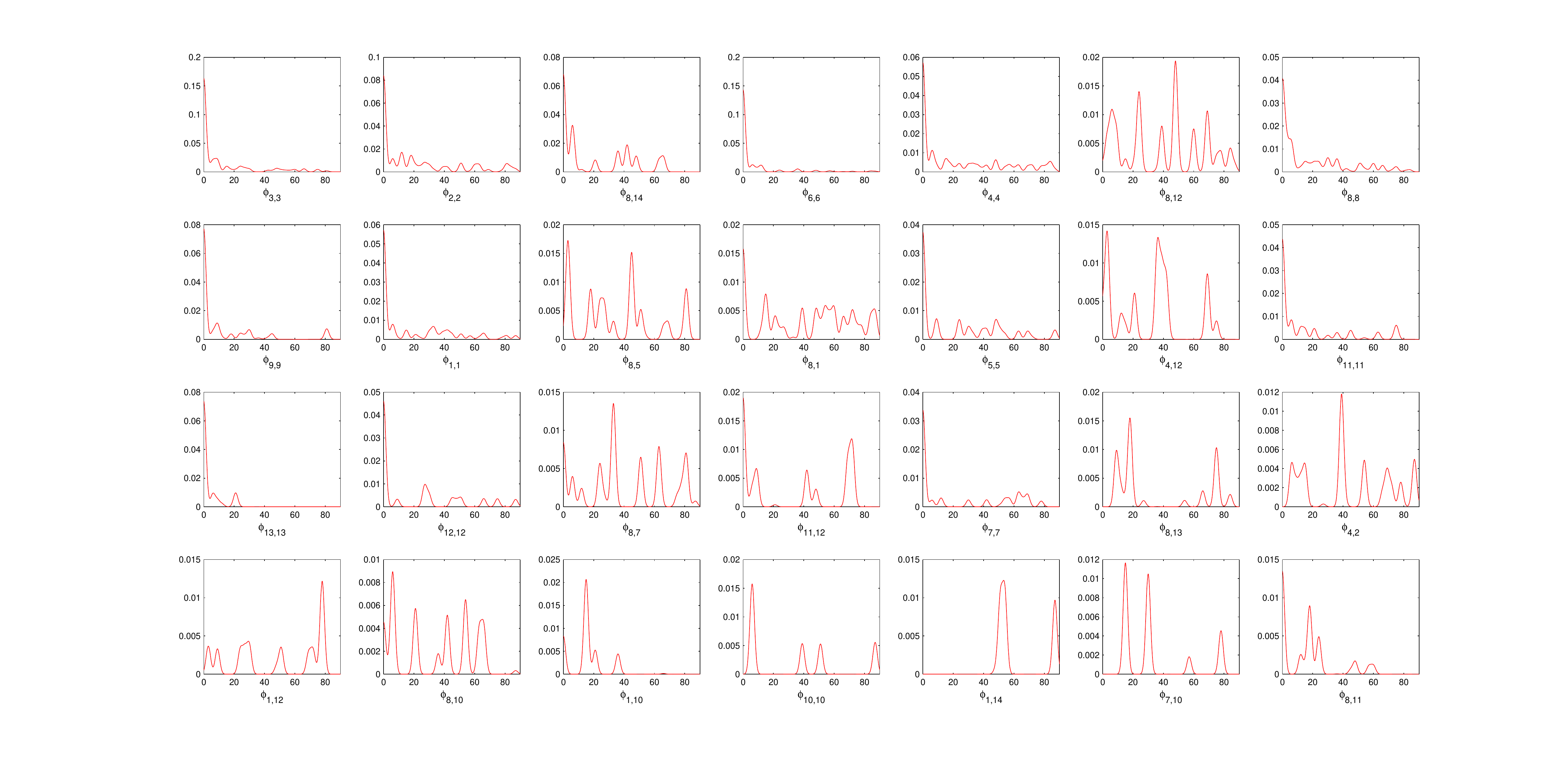}
\caption{The curves of the top-28 ${\phi}_{cc'}(t)$ between
different types of failures}
\end{center}
\end{figure}

First, depending on whether the impact functions have a delay part, we divide the trigger pattern of failures into two categories.
If the impact functions are close to 0 at the beginning, then we call these trigger patterns as delay trigger patterns.
Such as ${\phi _{8,13}}$ and  ${\phi _{7,10}} $ in figure 17 at the beginning, the value of this type of impact function is not very high ,even close to 0, but one or more spikes suddenly will appear over time, operation and maintenance personnel need to pay more attention to possible secondary faults at the corresponding moments where the spikes appear. From the perspective of the control field, the reason of this occurring phenomenon may be that there is delay between these systems, i.e. firefighting system to intake system and process system to control system.

In table 5, we will list the characteristics of delay trigger patterns, including the length of the delay and the moment when the peak risk occurs.

\begin{table}[]
\centering

\caption{The characteristics of delay trigger patterns}
\resizebox{\textwidth}{15mm}{
\begin{tabular}{lll}
\hline
Failure trigger     & Length of the delay(d) & Peak occurrence time(d) \\ \hline
Process system failure to Control system failure &  10    &      18.00    \\
Firefighting system failure to Intake system failure &  4    &      14.65    \\
Other failure to Gas Generator system failure &   45    &     53.20      \\ \hline
\end{tabular}}
\end{table}

Secondly, according to the attenuation of the impact function, we divide the rest
impact function into two categories. If the impact function can
decay to 10\% of its maximum value within 50 days, then we consider the trigger
pattern between the two failures is stable, otherwise, it is
unstable.we use ${\phi _{3,3}}$  and ${\phi _{2,2}}$ as examples, which is illustration in Figure 17. When a fault just occurs, the probability of a secondary fault is very high, but it gradually decreases with time.
The impact functions of this category are mainly the self-trigger impact function of the failures.
When these faults occur, the operation and maintenance personnel should remain vigilant within a short period of time after handling the fault to prevent occurring secondary faults, such as engine system failure, motor system failure, power supply system failure, control system failure, cooling system failure, gas generator system failure, Fuel gas system failure, compressor system failure, instrument air system failure, and Intake system failure. In table 6, we will list the higher risks duration of stable failure trigger.

\begin{table}[htbp]
    \centering
    \caption{Higher risks duration of stable failure trigger}
    \resizebox{\textwidth}{40mm}{\begin{tabular}{cc}
        \toprule  
        Failure trigger & Higher Risks duration(d) \\
        \midrule  
        Engine system failure to Engine system failure &0-10.0\\
        Motor system failure to Motor system failure&0-20.5\\
        Power supply system failure to Power supply system failure&0-48.7\\
        Lubricating oil system failure to Lubricating oil system failure&0-3.4\\
        Control system failure to Control system failure&0-37.2\\
        Gas Generator system failure to Gas Generator system failure&0-33.9\\
        Process system failure to Process system failure&0-22.1\\
        Instrument air system failure to Instrument air system failure&0-30.9\\
        \bottomrule  
    \end{tabular}}
\end{table}

More specifically, for a stable trigger pattern, after a failure
occurs, the probability of occurrence of a derivative fault is
generally reduced with time, while for the unstable trigger pattern,
the probability of occurrence of a derivative failure is only slightly
reduced, or instead will increase. Such as ${\phi _{8,12}}$ and ${\phi _{8,5}}$ depicting in Figure 17, this kind of impact function has a large value in a short time after the source failure occurs, and there are still several peaks appearing with the passage of time. After handling the source fault, the operation and maintenance personnel should not only be vigilant in a short period of time, but also should be vigilant at the moment of the corresponding peak value to prevent the occurrence of secondary faults, Such as instrument air system to control system and fuel gas system to control system. We will list the Peak appearance time of unstable failure trigger of in table 7.

\begin{table}[htbp]
    \centering
    \caption{Peak appearance time of unstable failure trigger}
    \resizebox{\textwidth}{70mm}{
    \begin{tabular}{cc}
        \toprule  
        Failure trigger & Peak occurrence time(d) \\
        \midrule  
        Other failure to Control system failure & 0.0\\
        Instrument air system failure to Control system failure&48.0\\
        Fuel gas system failure to Control system failure &3.0\\
        Cooling system failure to Cooling system failure& 0.0\\
        Gas Generator system failure to Control system failure &0.0\\
        Instrument air system failure to Power supply system failure&2.9\\
        Fuel gas system failure to Fuel gas system failure &0.0\\
        Intake system failure to Control system failure&33.0\\
        Instrument air system failure to Compressor system failure&0.0\\
        Process system failure to Control system failure&18.0\\
        Intake system failure to Intake system failure & 0.0\\
        Motor system failure to Power supply system failure&39.0\\
        Instrument air system failure to Gas Generator system failure&78.0\\
        Firefighting system failure to Control system failure&6.0\\
        Firefighting system failure to Gas Generator system failure&15.0\\
        Firefighting system failure to Firefighting system failure &6.0 \\
        Compressor system failure to Control system failure&0.0\\
        \bottomrule  
    \end{tabular}}
\end{table}

In summary, after we know the sequence of historical failure events, we can use the learned Weibull-Hawkes model to predict future failures, so as to respond to measures and reduce potential economic and security losses.

\section{Conclusion}

In this paper, we modify the structure of the Hawkes process, introduce the time-varying base intensity with the Weibull base intensity as an example, and propose a series of Hawkes process model. Based on our proposed model and EM-based algorithm, we provide an algorithm that effectively learns the parameters in the model. To demonstrate the effectiveness, stability and robustness of our model, we tested our proposed approaches both on time-varying base intensity data and constant base intensity data.

The most important point, which deserves our attention, is that in the practical application of the Hawkes process, assuming the base intensity is constant is extremely unrealistic. This assumption constricts the development of the Hawkes process in practical applications. Setting $h(t)$ is time-varying is more flexible and more generalization. This assumption is more in line with the needs of real-world applications. We learned our model on the historical data of the compressor station failure, got many valuable analyze results. After that, we made some suggestions for the actual production of the compressor station.
In the future, we are interested in the following work: create better parameter learning algorithm and devise new point process model that better fits the needs of actual production applications.

\section{Conflict of interest statement}

We declare that we have no financial and personal relationships with other people or organizations that can inappropriately influence our work, there is no professional or other personal interest of any nature or kind in any product, service and/or company that could be construed as influencing the position presented in, or the review of, the manuscript entitled. Survival Analysis of the Compressor Station Based on Hawkes Process with Weibull Base Intensity.

\section{Acknowledgment}

\bibliography{mybibfile}

\end{document}